\documentclass[acmtog]{acmart}
\acmSubmissionID{193}

\usepackage{booktabs} 

\usepackage[ruled]{algorithm2e} 

\SetAlFnt{\small}
\SetAlCapFnt{\small}
\SetAlCapNameFnt{\small}
\SetAlCapHSkip{0pt}

\usepackage{graphicx}
\usepackage{bm}
\usepackage{color}
\usepackage{multirow}
\usepackage{url}
\usepackage{float}
\usepackage{wrapfig}
\usepackage{subfigure}
\usepackage{colortbl}
\usepackage{xcolor}

\setcopyright{acmlicensed}
\acmJournal{TOG}
\acmYear{2023} 
\acmVolume{42} 
\acmNumber{4} 
\acmArticle{} 
\acmMonth{8} 
\acmPrice{15.00}
\acmDOI{10.1145/3592101}

\newcommand{\revision}[1]{{#1}}
\newcommand{\etal}{{et al.}}

\begin{document}
\title{AvatarReX: Real-time Expressive Full-body Avatars}

\author{Zerong Zheng}
\orcid{0000-0003-1339-2480}
\affiliation{%
 \institution{Department of Automation, Tsinghua University}
 \city{Beijing}
 \country{China}}
\email{zrzheng1995@foxmail.com}
\author{Xiaochen Zhao}
\orcid{0000-0001-8976-7723}
\affiliation{%
 \institution{Department of Automation, Tsinghua University and NNKosmos Technology}
 \city{Beijing}
 \country{China}
}
\email{zhaoxc19@mails.tsinghua.edu.cn}
\author{Hongwen Zhang}
\orcid{0000-0001-8633-4551}
\affiliation{%
 \institution{Department of Automation, Tsinghua University}
 \city{Beijing}
 \country{China}
}
\email{zhanghongwen@mail.tsinghua.edu.cn}
\author{Boning Liu}
\orcid{0000-0002-3014-6049}
\affiliation{%
 \institution{Department of Automation, Tsinghua University}
 \city{Beijing}
 \country{China}
}
\email{liuboning@mail.tsinghua.edu.cn}
\author{Yebin Liu}
\orcid{0000-0003-3215-0225}
\affiliation{%
 \institution{Department of Automation, Tsinghua University}
 \city{Beijing}
 \country{China}
}
\email{liuyebin@mail.tsinghua.edu.cn}

\begin{teaserfigure}
\centering
\includegraphics[width=1.0\linewidth]{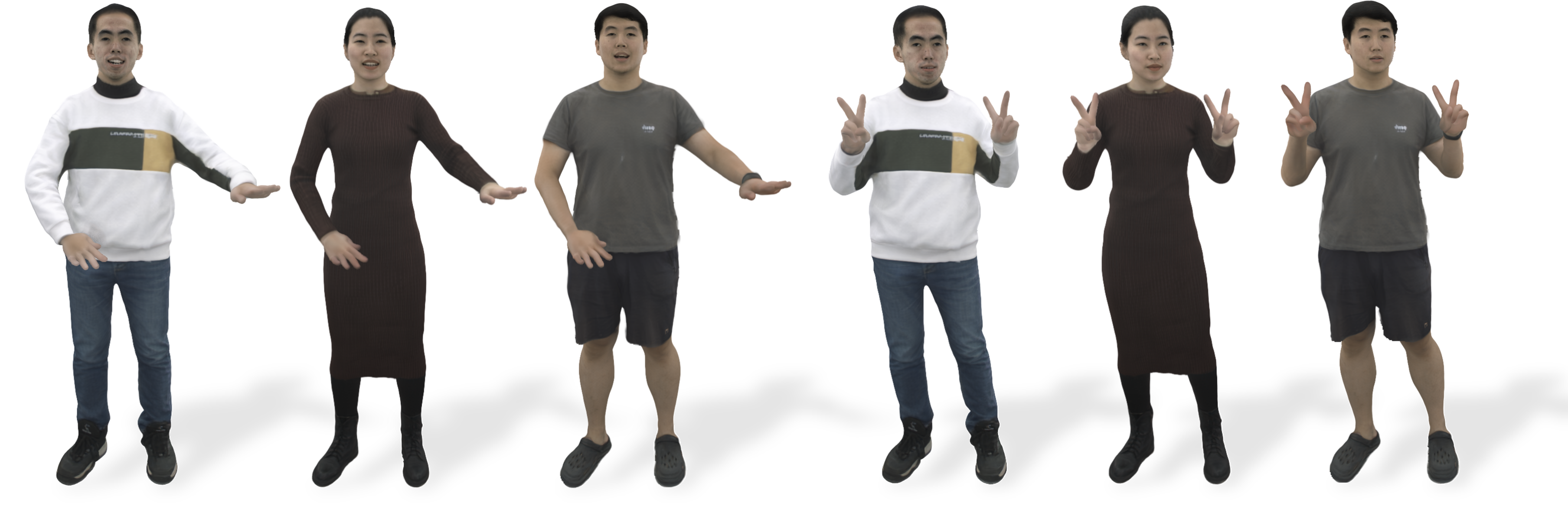}
\caption{\textbf{Example animation results produced by our method.} Our method can learn photorealistic full-body avatars that provides full controllability over the body pose, hand genstrue and the face expression all together. Moreover, it can be rendered in real time without compromising image quality.  }
\label{fig:teaser}
\end{teaserfigure}

\begin{abstract}
We present AvatarReX, a new method for learning NeRF-based full-body avatars from video data. The learnt avatar not only provides expressive control of the body, hands and the face together, but also supports real-time animation and rendering. To this end, we propose a compositional avatar representation, where the body, hands and the face are separately modeled in a way that the structural prior from parametric mesh templates is properly utilized without compromising representation flexibility. Furthermore, we disentangle the geometry and appearance for each part. With these technical designs, we propose a dedicated deferred rendering pipeline, which can be executed at a real-time framerate to synthesize high-quality free-view images. The disentanglement of geometry and appearance also allows us to design a two-pass training strategy that combines volume rendering and surface rendering for network training. In this way, patch-level supervision can be applied to force the network to learn sharp appearance details on the basis of geometry estimation. Overall, our method enables automatic construction of expressive full-body avatars with real-time rendering capability, and can generate photo-realistic images with dynamic details for novel body motions and facial expressions. 

\end{abstract}

%
%

\ccsdesc[500]{Computing methodologies~Computer vision}
\ccsdesc[500]{Computing methodologies~Rendering}
%
%

\keywords{full-body avatars, real-time rendering, neural radiance fields}

\maketitle

\section{Introduction}
\label{sec:intro}

Animatable human avatar modeling, as an important topic in special effects industry, can be applied in many applications such as content creation and immersive entertainment. Virtual characters are believed to have the potential to open up a new way for people to interact with others or intelligent machines in AR/VR settings~\cite{ChuMTFS20ModularCodecAvatar}. Unfortunately, the traditional pipeline for  creating 3D human avatars involves tedious procedures, including scanning, meshing, rigging and many more. Furthermore, such a pipeline requires expert knowledge and sophisticated capture systems, limiting its access and increasing its cost~\cite{Alexander2010DigitalEmily}.

In order to lower the entry barrier for novices and automate the workflow of experienced artists, researchers have devoted great efforts in learning 3D human body avatars from images or videos of real humans. Building upon explicit meshes~\cite{timur2021driving_signal,habermann2021realtimeDDC,Xiang2021ModelingClothing}, implicit radiance fields~\cite{peng2021animatable_nerf,zheng2022structured,Wang2022ARAH,Su2021ANeRF,weng_humannerf_2022_cvpr} or both~\cite{Lombardi2021MVP,neural_actors,Remelli2022TexelAligned}, current approaches are able to synthesize realistic human motions under free-view points. Despite the plethora of these systems, most of them only model the torsos and the limbs, \revision{without combining} other fine-grained body parts like faces and hands. 
However, an expressive human avatar demands for full controllability of the body, hands and the face together, as the essential nuance of human behaviors is conveyed through a concert of body movements, hand gestures and facial expressions. 
Up to now, only research works from industrial labs are able to achieve this goal~\cite{timur2021driving_signal,Remelli2022TexelAligned,Xiang2021ModelingClothing,Xiang2022DressingAvatars}. Unfortunately, their methods rely on video data captured from dense-view camera rigs, which are not accessible for most individuals and academic organizations.

Apart from expressiveness, another unsolved challenge lies in the rendering speed. 
In many applications, the avatars are expected to interact with users as real human-beings, which emphasizes the need for 
real-time animation and rendering. 
However, the most recent approaches~\cite{peng2021animatable_nerf,zheng2022structured,Wang2022ARAH,Su2021ANeRF,neural_actors} in this field are typically built upon neural radiance fields (NeRF)~\cite{mildenhall2020nerf}, which densely samples the 3D space and queries the network millions of times for volume rendering. Consequently, they are difficult to render a dynamic avatar \revision{at a real-time framerate}, preventing them from being adopted in interactive scenarios.  

In this work, we present \textbf{Re}al-time e\textbf{X}pressive \textbf{Avatar} (AvatarReX), a novel system that simultaneously achieves expressive control and real-time animation of full-body human avatars.
To this end, we propose a compositional representation that models the face, hands and the body with independent implicit fields (Section~\ref{sec:avatar_reprez}). 
Such a compositional design allows us to adopt the most suitable technique for each part according to its characteristics in shape and texture variations. 
In our method, all part representations rely on the corresponding parametric templates, 
\textit{i.e.}, SMPL-X~\cite{SMPL-X:2019} for the body, MANO~\cite{Romero2017MANO} for the hands and Faceverse~\cite{wang2022faceverse} for the face, 
but the prior in these templates is leveraged in totally different manners. 
For example, we employ structured local radiance fields~\cite{zheng2022structured} for clothed body representation in order to eliminate the reliance on the SMPL-X topology, and, \revision{in contrast}, directly build the the radiance fields of hands on top of the MANO geometry model. This is because clothes vary in topology, while the shape variation of hands is rather limited. 
To enhance the expressiveness of our avatar, we introduce several novel techniques for the body and face parts.
Specifically, to capture the rich dynamic details like cloth wrinkles exhibited in the body part, we propose explicit feature patches to facilitate the learning of these appearance details. 
Meanwhile, as the face part contains many subtle yet important details relating to different expressions, we utilize convolution networks to extract distinctive features from the expression space of Faceverse, which is a better condition for high-fidelity facial rendering.
Overall, each part representation in our avatar is carefully designed, ensuring that the 3D prior of the parametric templates is properly utilized without sacrificing representation power and model flexibility.

With these building blocks at hand, we can now derive the final expressive avatar by assembling the face, hands and the body into one holistic model. However, it remains a challenge to train an avatar with sharp appearance details and render it efficiently at test time.  
To resolve these problems, we further improve our avatar representation by disentangling the geometry and appearance for each part and using the signed distance function (SDF) as the common geometric representation. 
These technical designs enable us to develop a real-time rendering pipeline based on deferred rendering (Section~\ref{sec:realtime}). 
The core of our pipeline is to take advantage of the implicit surface definition in SDF for surface rendering. Compared to volume rendering in most NeRF-based methods~\cite{mildenhall2020nerf}, surface rendering avoids the need for expensive sampling along camera rays, leading to a significant speedup. 
We further accelerate this process with a dedicated deferred shading scheme, where the geometry model is firstly reconstructed in the form of an explicit SDF volume, enabling fast surface location for the later color evaluation step. 
As a result, our method successfully accelerates the avatar rendering process by two orders of magnitude without compromising the quality. 

In addition to testing acceleration, the disentanglement of geometry and appearance can also benefit network training. 
To this end, we combine deferred surface rendering with volume rendering to form a two-pass training strategy (Section~\ref{sec:training}). On one hand, volume rendering allows the geometry networks to learn various cloth shape from scratch with sparse pixel supervision. On the other, when the surface reconstruction is available, surface rendering minimizes network queries, making it possible to apply perceptual supervision on image patches, which is essential to force the color network to learn high-quality appearance. 
With our disentangled design, we can combine the merits from both worlds into two training passes, and consequently obtain  an avatar with sharp appearance details.





In summary, our system is able to create expressive full-body avatars with high-quality appearance details, and the avatars can be animated and rendered in real time.
\revision{
The training data for our avatar is multi-view videos of approximately 2000 frames in length, captured from 22 cameras (16 for the body and hands and 6 for the face). 
After data collection, our method can automatically learn the avatar representation without the need for pre-scanning efforts or other manual intervention. Moreover, the learned avatar can be driven in real-time given new body poses, hand gestures and facial expressions. 
}
Experimental results clearly demonstrate the potential of applying our method in interactive applications.

\section{Related Work}

\textbf{Body Avatars.} 
In the last decade, many efforts have been made to achieve animatable human avatars. 
Pioneer works in this field resort to statistical mesh templates~\cite{loper2015smpl,STAR:2020,SMPL-X:2019,TotalCapture2018} to model minimally clothed bodies.
To handle the varying shape of clothing, recent methods explore more flexible representations such as implicit fields to model the shapes of clothed humans~\cite{tiwari21neuralgif,LEAP:CVPR:21,Saito:SCANimate:2021,chen2021snarf,li2022avatarcap,zhang2023closet,lin2022fite}.
For instance, Neural-GIF~\cite{tiwari21neuralgif} factorizes human motion into articulation and non-rigid deformation, and learns to map every point in the space to a canonical pose using backward skinning.
LEAP~\cite{LEAP:CVPR:21} and SCANimate~\cite{Saito:SCANimate:2021} learn the forward and backward skinning fields with neural networks and regularize the cycle consistency between them. 
For better generalization to unseen poses, SNARF~\cite{chen2021snarf} proposes a differentiable forward skinning model based on iterative root finding, which finds the canonical correspondences of any query point in the posed space. 

To acquire animatable characters with color, traditional pipelines typically reconstruct a subject-specific textured mesh in advance, and then generate its motions using physics simulation~\cite{Stoll2010Videobased,DRAPE2012}, database retrieval and blending~\cite{Xu2011Videobased}, or deformation space modeling~\cite{TotalCapture2018,habermann2021realtimeDDC}. 
With accurate tracking of the underlying geometry, Bagautdinov \etal~\shortcite{timur2021driving_signal} model high-fidelity avatars by decoding dynamic geometry and appearance from disentangled driving signals.
This approach is further extended in \cite{Xiang2021ModelingClothing,Xiang2022DressingAvatars} by representing the clothing as a separate layer in order to recover sharper garment boundaries.
Their reliance on pre-scanning subject-specific templates can be eliminated via deforming a general body template.
For instance, several works proposed to directly learn this deformation from geometric data~\cite{CAPE:CVPR:20,pons2017clothcap,Ma:POP:2021} or RGB videos~\cite{alldieck2018videoavatar,alldieck2018videoavatar_detailed,burov2021dsfn}. 
\revision{
The texture map and the rasterization step in these methods are later replaced with neural texture maps and image decoders~\cite{Liu2018Neural,liu2020NeuralHumanRendering,Shysheya2019TNR,raj2020anr,Prokudin2021SMPLpix,HVTR:3DV2022} to achieve more photo-realistic rendering. 
}

In the past three years, neural volumetric representations have demonstrated impressive results on novel view synthesis of static scenes~\cite{mildenhall2020nerf} or dynamic scenes~\cite{Lombardi:2019:NeuralVolumes}. Since then, great efforts have been made to extend these neural representations to human avatars. For example, Neural Body~\cite{peng2021neuralbody} uses SMPL~\cite{loper2015smpl} to establish correspondences across different frames and trains a sparse convolution network to convert the SMPL vertices into a radiance volume. It supports high-quality view synthesis, but can only playback the training sequences. To achieve better pose generalization, recent works adopt a disentangled representation, which maps the points in different poses to a canonical radiance field using inverse skinning~\cite{peng2021animatable_nerf,weng_humannerf_2022_cvpr,Li2022TAVA,Wang2022ARAH,Su2021ANeRF,li2023posevocab}.
For modeling the dynamic appearance details, Neural Actors~\cite{neural_actors} introduces a texture map as an additional condition.  In contrast to them, Zheng \etal~\shortcite{zheng2022structured} present a structured local representation, where the radiance field of dynamic characters is assembled by a set of local ones, similarly to MVP~\cite{Lombardi2021MVP}. 
\revision{
DANBO~\cite{su2022danbo} employs a part-based volumetric representation defined by the skeleton structure using graph neural networks. 
}
Recently, Remelli \etal~\shortcite{Remelli2022TexelAligned} combine localized volumetric primitives with the dense signal from image observations, allowing faithful synthesis of appearance details like cloth wrinkles. 
\revision{
However, their avatars can only be driven by the same person in the same attire due to its requirement of driving views. In contrast,  our method takes as input solely the pose parameters and the expression coefficients no matter where they come from, thus our avatar can be driven by another person or other signal sources. 
}
Concurrent to us, TotalSelfScan~\cite{dong2022totalselfscan} reconstructs a full-body model from self-portrait videos of faces, hands, and bodies, but it can only produce articulated body motions and fails to synthesize photo-realistic dynamic appearance.


\begin{figure*}
    \centering
    \includegraphics[width=0.98\linewidth]{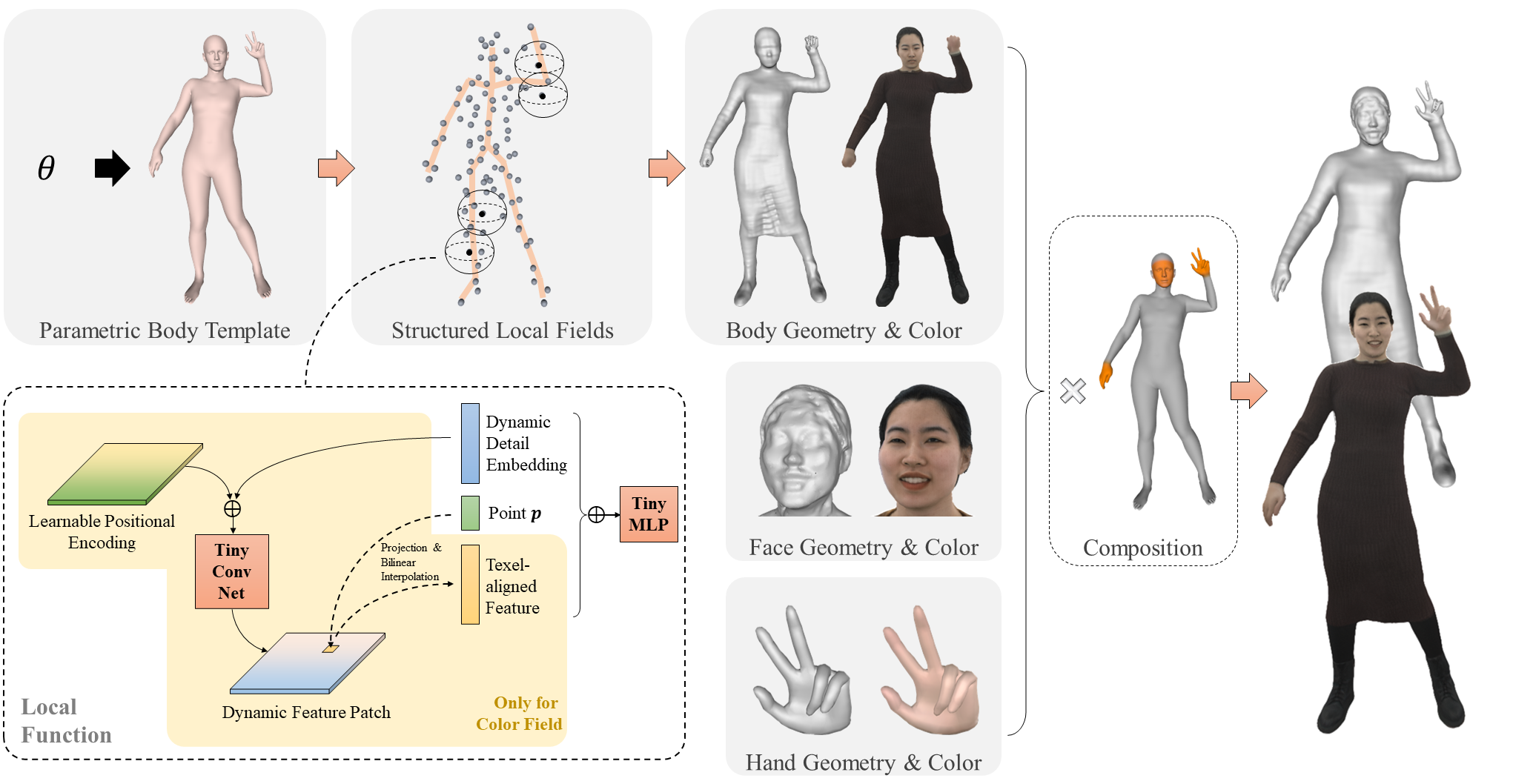}
    \caption{\textbf{Illustration of our compositional avatar representation.} Our expressive avatar is composed of three parts, namely the major body, the hands and the face. For clarity, we only illustrate the body representation here, and leave the other two in Figure~\ref{fig:avatar_reprez:hand} and Figure~\ref{fig:avatar_reprez:face}. The core of our body representation is a set of structured local implicit fields, and we enhance their detail representation power by introducing an explicit dynamic feture patch for each field. }
    \label{fig:avatar_reprez:body}
\end{figure*}

\textbf{Face Avatars.}
Similar to human bodies, face avatar techniques have also undergone significant advancements. The seminal work dating back to 1999 built the first 3D morphable model, enabling  representation of  facial shapes by embedding different identities and expressions into low-dimensional PCA spaces~\cite{BlanzV99:3Dmm}. 
To model complex deformations and textures, researchers have exploited more advanced modeling tools, such as multi-linear models~\cite{VlasicBPP06,CaoWZTZ14}, nonlinear models~\cite{GuoCZ21,Tran018Nonlinear} and the articulated control of expression~\cite{FLAME:SiggraphAsia2017}. Recent methods further recover high-frequency deformations of expressions by learning additional displacement maps on top of the base mesh model~\cite{FengFBB21Learning,DanecekBB22EMOCA,grassal2022NHA}. Moreover, some researchers propose reconstructing facial avatars with remarkable quality for immersive telepresence based on dense multiview capture systems~\cite{LombardiSSS18DeepAppearanceModels,ChuMTFS20ModularCodecAvatar,MaSSWLTS21PixelCodecAvatar,wang2023styleavatar}. 

Since the debut of implicit representations like DeepSDF~\cite{park2019deepsdf} or NeRF~\cite{mildenhall2020nerf}, it becomes an growing trend to model 3D faces or heads in an implicit fashion. Building upon neural implicit functions, Yenamandra \etal~\shortcite{yenamandra2021i3dmm} developed the first deep implicit 3D morphable model of full heads including faces and hairs. Similarly, Hong \etal~\shortcite{hong2021headnerf} adopt NeRF to create a parametric head model that supports  high-fidelity head image rendering in real-time. However, these models focus on learning generic head models using data from multiple subjects, often lacking personalized appearance details. To address this issue, NeRFace~\cite{gafni2021nerface} proposes a personalized head avatar by taking expression coefficients as the additional inputs for the head NeRF, and demonstrates state-of-the-art reenactment and rendering results. IMAvatar~\cite{ZhengABCBH22IMAvatar} incorporates skinning fields with an implicit morphing-based model, which allows better geometry reconstruction and stronger generalization capability for novel expressions. Gao \etal~\shortcite{Gao2022nerfblendshape} bridge traditional mesh blendshapes with voxel-based implicit fields, enabling fast construction of personalized head NeRF models from monocular videos. 
\revision{
These methods typically require dense images or videos as input, and efforts have been made to alleviate the reliance on large amounts of data~\shortcite{Cao2022Authentic,Zhang2022FDNeRF,xu2023latentavatar,xu2023avatarmav}. 
}
Some recent research works also propose to replace the expression coefficients with other driving signals, such as audio~\cite{Guo2021ADNeRF} and gaze~\cite{Richard2021AudioGaze}.


\textbf{NeRF Acceleration.} 
Numerous works have emerged with the purpose of speeding up static NeRF using explicit data structures including feature maps, voxels and tensors. 
For instance, DVGO~\cite{DVGO-na63} achieves fast convergence through an explicit representation of a density voxel grid and a feature voxel grid. 
Plenoxels~\cite{Plenoxels2022-na57} and PlenOctree~\cite{Plenoctrees2021-nfds81} model a scene through a hierarchical 3D grid with spherical harmonics, which can realize an optimization with two orders of magnitude faster than NeRF. 
DIVeR~\cite{DIVeR2022-na75} accelerates volumetric rendering by limiting ray marching to a fixed number of hits on the voxel grid.
Hashing encoding~\cite{InstantNGP2022-na44} and tensor decomposition~\cite{TensoRF2022-na9} are also used as compact representations for NeRF acceleration. 
Similar techniques have also been applied for dynamic scene rendering. For example, DeVRF~\cite{DeVRF2022-nfds32} enables fast non-rigid neural rendering with both 3D volumetric and 4D voxel fields. 
TiNeuVox~\cite{TiNeuVox2022-na13} represents scenes with optimizable explicit data structures and accelerates radiance fields modeling, while Wang \etal~\shortcite{Fourierplenoctrees2022-nfds72} extended PlenOctree into free-view video rendering. Despite demonstrating significant speedup in NeRF training and testing, most of these works can only render static scenes or playback a dynamic sequence, making them unsuitable for character animation settings.

In this paper, we propose a novel real-time rendering pipeline based on deferred surface rendering. The philosophy behind our design is similar to MobileNeRF~\cite{Chen2022MobileNeRF}, which also computes the pixel color in the image space rather than volume rendering. However, MobileNeRF only works for static scenes because it represents the scene as a fixed triangle mesh. In contrast, the topology of our avatar varies from pose to pose, and cannot be modeled with a stationary mesh. Therefore, we extract the pose-dependent geometry model on the fly via taking advantage of the SDF definition and the parallelism in our representation. 
We further disentangle geometry and appearance to reduce the computational burden in geometry reconstruction. Although disentangling geometry and appearance has been proposed for multi-view geometry reconstruction~\cite{yariv2020idr}, we are the first to employ it for real-time rendering. This requires us to accelerate the expensive operation of sphere tracing, and we achieve this goal by caching SDF values in an explicit volumetric grid.

\section{Avatar Representation}
\label{sec:avatar_reprez}


This section discusses how we represent our expressive avatar. The avatar is composed of three parts, namely the body, the hands and the face. Considering their different characteristics in shape and texture variations, we design different neural implicit representations for them. They all rely on the corresponding parametric mesh templates, \textit{i.e.}, SMPL-X for the body~\cite{SMPL-X:2019}, MANO for the hands~\cite{Romero2017MANO} and Faceverse for the face~\cite{wang2022faceverse}, but these templates is leveraged in totally different manners, as we will discuss in Section~\ref{sec:avatar_reprez:body}, \ref{sec:avatar_reprez:hand} and \ref{sec:avatar_reprez:face}. Finally, we introduce the technique for combining all the parts into one final avatar model in Section~\ref{sec:avatar_reprez:composition}.

\textbf{Notation.} In the following text, we denote the part representation with $\mathcal{A}_{*}(s)$, where $s$ is the driving signal and the subscript ``$*$'' can be $\{\text{``}\mathrm{B}\text{''}, \text{``}\mathrm{H}\text{''}, \text{``}\mathrm{F}\text{''}\}$, representing ``body'', ``hand'' and ``face'', respectively. 
Accordingly, the driving signal $s$ comes from body poses $\bm{\theta}$, hand poses $\bm{\phi}$ or facial expressions $\bm{\psi}$.  
Each part representation consists of two components, namely a geometry field $\mathcal{G}_{*}$ and a view-dependent color field $\mathcal{C}_{*}$:  
\begin{equation}
    \mathcal{A}_{*}(s) = \{ \mathcal{G}_{*}(\bm{p} | s), \mathcal{C}_{*}(\bm{p}, \bm{v} | s) \}, 
\end{equation}
where $\bm{p}$ and $\bm{v}$ are the 3D point position and the viewing direction, respectively. 
For ease of notation, we drop the dependency on view directions when discussing the color fields in the upcoming sections. 

In our method, we represent the geometry field as a signed distance function (SDF), where the true surface is embedded as its zero-level set $\{ \bm{p}\in\mathbb{R}^3 | \mathcal{G}_{*}(\bm{p} | s) = 0  \}$. 
Note that we do not follow the vanilla NeRF~\cite{mildenhall2020nerf} and most NeRF-based avatar works~\cite{peng2021animatable_nerf,zheng2022structured} that use one network to simultaneously model the geometry (density) field and the color field. Instead, we model them in a disentangled fashion and use smaller network size for the geometry fields. 
Such a design is for the purpose of real-time implementation as well as two-pass training, as we will discussed in Section~\ref{sec:realtime} \& \ref{sec:training} .

\subsection{Body Representation}
\label{sec:avatar_reprez:body}
We adopt structured local radiance fields~\cite{zheng2022structured} as the representation of our body geometry field. Here we briefly review its construction for completeness. 
Specifically, we pre-define a set of nodes $\{\bm{\bar{n}}_i\}_{i=1}^N$ on the SMPL-X model~\cite{SMPL-X:2019} via farthest point sampling. Since the nodes are sampled from the SMPL-X model, each of them has associated skinning weights, thus can be driven by the skeleton. Furthermore, we allow the nodes to have their own residual movements  $\Delta\bm{n}_i$ to represent the non-rigid deformation of garments.

Around each node, we construct a local implicit field centered at it. Take the geometry field $\mathcal{G}_\mathrm{B}$ as an example. For each node, we define a function $\mathcal{G}_{i}$ in a local space around it, and use a tiny MLP to represent this function. 
This MLP takes as input a coordinate in the local space of node $i$ and outputs a high-dimensional feature vector, which will be blended with the outputs from other local MLPs and finally decoded into an SDF value. 
Formally, given any point $\bm{p}\in\mathbb{R}^3$ in the posed space of pose $\bm{\theta}$, we first calculate its coordinate in the local space of node $i$ as:
\begin{equation}
    \bm{p}_i = \mathbf{T}^{-1} \bm{p} - \left(\bm{\bar{n}}_i + \Delta\bm{n}_i\right), 
\end{equation}
where $\mathbf{T}$ is the skinning matrix computed from the pose parameter $\bm{\theta}$ using linear blending skinning and $\bm{n}_i$ is the position of node $i$ in the posed space. 
After that, we feed it into the local network $\mathcal{G}_i$ and blend the feature vectors produced by all the local MLPs:
\begin{equation}
\label{eqn:avatar_reprez:body:feat_fusion}
    \bm{f} = \frac{\sum w_i \mathcal{G}_i(\bm{p}_i, \bm{e}_i)}{\sum w_i}, 
\end{equation}
where $\bm{e}_i$ is a dynamic detail embedding predicted from the pose parameters and models the fine-grain deformations that cannot be represented by node movements.
The blending weight $w_i$ is calculated as:
\begin{equation}
\label{eqn:avatar_reprez:body:weight}
    w_i = \max\left\{\exp\left( \frac{-\Vert\bm{p} - \bm{n}_i\Vert_2^2}{2\sigma^2}\right) - \epsilon, 0\right\}, 
\end{equation}
where $\sigma$ and $\epsilon$ are hyperparameters controlling the influence radius of the local networks. 
The blended feature $\bm{f}$ is fed into an additional MLP $\mathcal{G}_\mathrm{blending}$ to compute the SDF value of $\bm{p}_i$:
\begin{equation}
    \mathcal{G}_\mathrm{B}(\bm{p} | \bm{\theta}) = \mathcal{G}_\mathrm{blending} (\bm{f}).
\end{equation}

\textbf{Dynamic Feature Patch.} 
The color field of our body representation can be modeled in a similar way to the geometry field, \textit{i.e.}, with a set of local MLPs $\{ \mathcal{C}_i \}$ and a blending MLP $\mathcal{C}_\mathrm{blending}$. 
However, due to the low-frequency bias~\cite{tancik2020fourfeat}, the local MLPs $\{ \mathcal{C}_i \}$ are not powerful enough to represent the high-frequency details like the cloth wrinkles and texture patterns. 
To address this limitation, we introduce a dynamic feature patch for each local function of the body color field. 
Compared to purely implicit representation, an explicit feature patch contains more spatial information and provides stronger capability to store the information about high-frequency local details~\cite{neural_actors}. 
The feature patch, denoted as $\mathbf{F}_i$, is regressed by a tiny convolution network from the dynamic detail embedding $\bm{e}_i$ and a 2D learnable positional encoding, as illustrated in Figure~\ref{fig:avatar_reprez:body}. 
Given the local point position $\bm{p}_i$, we project it onto the feature patch: 
\begin{equation}
    \bm{u}_i = \Pi_i \bm{p}_i, 
\end{equation}
where $\Pi_i\in\mathbb{R}^{2\times3}$ is the projection matrix used for projecting 3D points along a specific direction. Here we pre-compute the projecting direction by averaging the normal orientations of the SMPL-X vertices that locate nearby node $i$. After that, we query for the feature vector at $\bm{u}_i$ through bilinear interpolation:
\begin{equation}
    \bm{f}_{i} = \mathrm{Bilinear}(\mathbf{F}_i, \bm{u}_i). 
\end{equation}
The interpolated feature vector is finally taken as an additional input by the local function $\mathcal{C}_i$ to produce the final color value:
\begin{equation}
\begin{split}
    &\bm{f}^{'} = \frac{\sum w_i \mathcal{C}_i(\bm{p}_i, \bm{e}_{i}, \bm{f}_{i})}{\sum w_i},  \\  
    &\mathcal{C}_\mathrm{B}(\bm{p} | \bm{\theta}) = \mathcal{C}_\mathrm{blending} (\bm{f}^{'}).    
\end{split}
\end{equation}
As shown in Section~\ref{sec:exp:ablation}, the proposed dynamic feature patches allow our color networks to learn more appearance details for the major body. 


\begin{figure}
    \centering
    \includegraphics[width=1.0\linewidth]{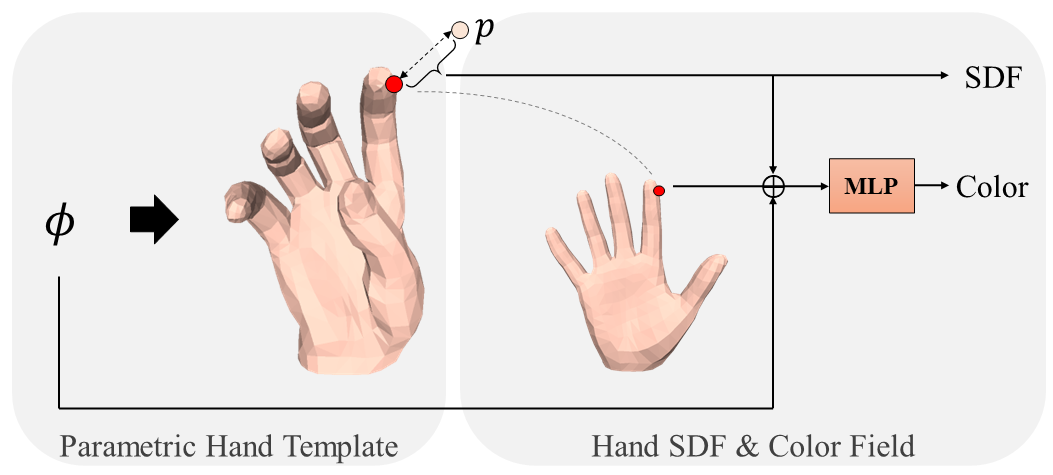}
    \caption{\textbf{Hand representation.} We directly use the SDF from the parametric template as the geometry field of our hand representation, and learn an MLP to model the color field in the canonical space. }
    \label{fig:avatar_reprez:hand}
\end{figure}

\subsection{Hand Representation}
\label{sec:avatar_reprez:hand}

Unlike the major body with clothing, hands show limited variations in shape and topology. Therefore, we directly use the surface of a parametric hand template to construct the geometry field of the hand, thus alleviating the need for learning complex articulated motion of fingers. Here we use MANO~\cite{Romero2017MANO} as the base hand representation. 
For any spatial point $\bm{p}\in\mathbb{R}^3$ in hand pose $\bm{\phi}$, we project it onto the MANO mesh by first calculating its barycentric projection on each triangle face of the MANO surface and then finding the nearest one. Mathematically, this procedure can be formulated as:
\begin{equation}
\label{eqn:avatar_reprez:hand:sdf}
\begin{split}
    (u^*, v^*, w^*, \mathbf{F}^*) = \arg&\min_{u, v, w, \mathbf{F}} ||\bm{p} - \mathrm{Barycentric}(\mathbf{F}, u, v, w)||^2_2, \\
    s.t., \quad  &0\leq u, v, w \leq 1, \\ 
    &u+v+w=1
\end{split}
\end{equation}
where $\mathbf{F}$ is a triangle of the MANO mesh and $\mathrm{Barycentric}(\ldots)$ is the barycentric interpolation function. Using the barycentric coordinates $(u^*, v^*, w^*)$ and the interpolation function, we determine the nearest point of $\bm{p}$ on the MANO mesh, which we denoted as $\bm{m}_{\mathrm{np}}$. Similarly, we can compute its normal direction $\bm{n}_{\mathrm{np}}$ through barycentric interpolation. The SDF value of $\bm{p}$ can be finally calculated as:
\begin{equation}
     \revision{\mathcal{G}_\mathrm{H}(\bm{p} | \bm{\phi}) = } \left\{ 
     \begin{array}{cl}
     ||\bm{p} - \bm{m}_{\mathrm{np}}||_2, & \bm{n}_{\mathrm{np}}^{\top}\left(\bm{p} - \bm{m}_{\mathrm{np}}\right) \geq 0 \\
     -||\bm{p} - \bm{m}_{\mathrm{np}}||_2, & \bm{n}_{\mathrm{np}}^{\top}\left(\bm{p} - \bm{m}_{\mathrm{np}}\right) < 0 \\
     \end{array}
     \right. .
\end{equation}

To model the color field of the hand, we turn to a neural perspective. We first calculate the canonical position of $\bm{m}_{\mathrm{np}}$ via interpolating the position of $\mathbf{F}$ in the canonical pose according to the barycentric coordinate. The interpolated result $\bar{\bm{m}}_{\mathrm{np}}$, together with the normal direction in posed space $\bm{n}_{\mathrm{np}}$, the signed distance $\mathrm{SDF}(\bm{p})$ and the hand pose $\bm{\phi}$, is fed into an MLP to produce the color value:
\begin{equation}
    \mathcal{C}_\mathrm{H}(\bm{p} | \bm{\phi}) = \mathcal{C}^{'}_\mathrm{H} \left(\bar{\bm{m}}_{\mathrm{np}}, \bm{n}_{\mathrm{np}}, \mathrm{SDF}(\bm{p}), \bm{\phi}\right), 
\end{equation}
where $\mathcal{C}^{'}_\mathrm{H}$ denotes the MLP network. 
In this way, our method is able to model the pose-dependent appearance of hands without the burden of learning complex hand motions. Figure~\ref{fig:avatar_reprez:hand} illustrates our hand representation.

\subsection{Face Representation}
\label{sec:avatar_reprez:face}

Compared to the body and the hands, representing the face is more challenging as humans are social animals that rely on facial expressions to read and convey emotions. As a result, we need to achieve both photo-realistic synthesis and accurate expression control in order to overcome the well-known uncanny valley. 
To this end, we propose to combine the the rendering power of NeRF and the prior knowledge from the facial morphable model~\cite{wang2022faceverse}. 

\begin{figure}
    \centering
    \includegraphics[width=1.0\linewidth]{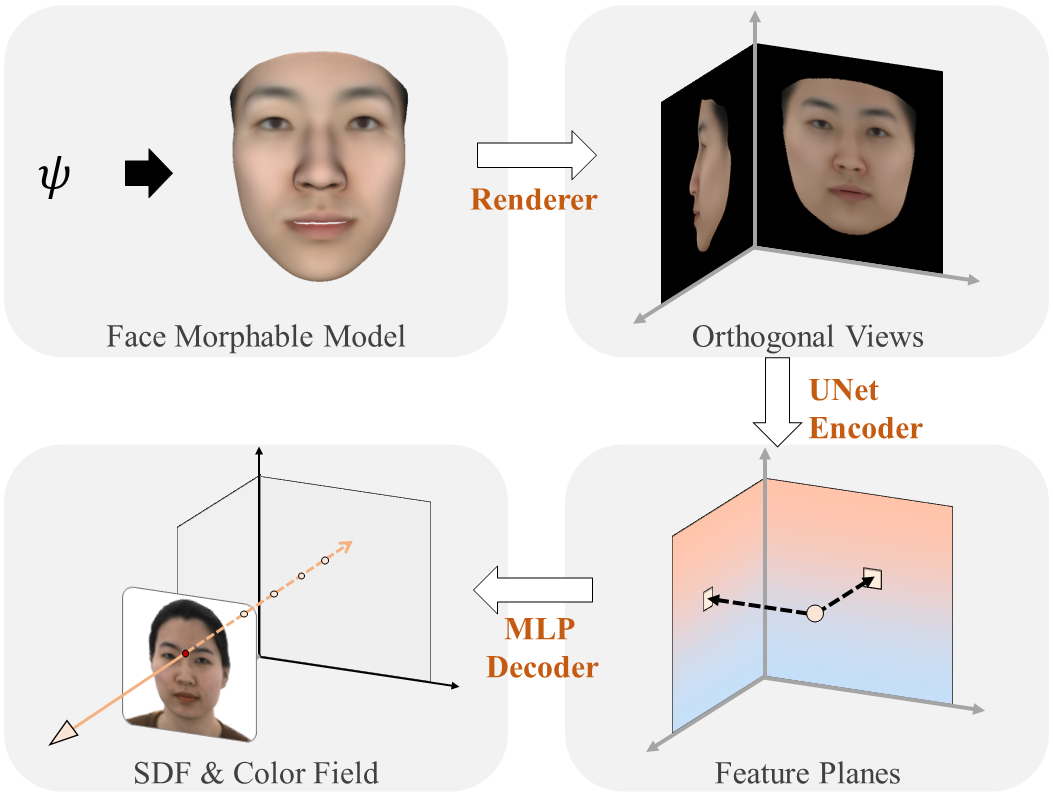}
    \caption{\textbf{Face representation.} We condition NeRF on the orthogonal views of a 3D morphable model, which provides structural prior of the face and controllability over expressions. }
    \label{fig:avatar_reprez:face}
\end{figure}

Given the expression coefficients $\bm{\psi}$, we first compute the corresponding 3D facial model. The model is an extremely coarse approximation of the real face, but it provides structural information about the specific expression. 
To utilize this structural knowledge in an efficient manner, we take inspiration from EG3D~\cite{Chan2021EG3D} and propose to learn a triplane-alike facial avatar representation. 
Specifically, we render the model from its front view and two side views using orthogonal projection, as illustrated in Figure~\ref{fig:avatar_reprez:face}. The rendered images are passed through three convolutional neural encoders, which extract the corresponding feature tri-planes denoted as $\{ \mathbf{F}_\mathrm{front}, \mathbf{F}_\mathrm{left}, \mathbf{F}_\mathrm{right} \}$. Given a spatial point $\bm{p}\in\mathbb{R}^3$, we project it onto the feature tri-planes and retrieve its pixel-aligned feature vectors through bilinear interpolation:
\begin{equation}
    \bm{f}_{v} = \mathrm{Bilinear}(\mathbf{F}_v, \Pi_v(\bm{p})), 
\end{equation}
where $\Pi_v\in\mathbb{R}^{2\times3}$ is the projection matrix and the subscript $v\in\{ \text{``}\mathrm{front}\text{''}, \text{``}\mathrm{left}\text{''}, \text{``}\mathrm{right}\text{''} \}$ denotes the projection direction. The feature vectors, together with the point coordinate, are fed into an MLP to produce the sign distance of $\bm{p}$:
\begin{equation}
 \mathcal{G}_\mathrm{F}(\bm{p} | \bm{\psi}) = \mathcal{G}_\mathrm{F}^{'} (\bm{p}, \bm{f}_\mathrm{front}, \bm{f}_\mathrm{left}, \bm{f}_\mathrm{right}), 
\end{equation}
where $\mathcal{G}_\mathrm{F}^{'}$ is an MLP network. The color of $\bm{p}$ is predicted in a similar way using another set of convolutional encoders and another MLP. 

Previous works on NeRF-based facial avatar typically use pure MLPs as the network architecture and take the global expression coefficients as the network condition~\cite{gafni2021nerface,ZhengABCBH22IMAvatar}. 
Compared to them, the explicit feature tri-planes allow us to keep the MLP decoder as light-weight as possible, thus reducing the computational cost of neural rendering. Furthermore, the feature tri-planes provide spatially varying conditioning for the 3D space, which has stronger power in representing appearance details than a global expression condition.  

\begin{figure*}
    \centering
    \includegraphics[width=1.0\linewidth]{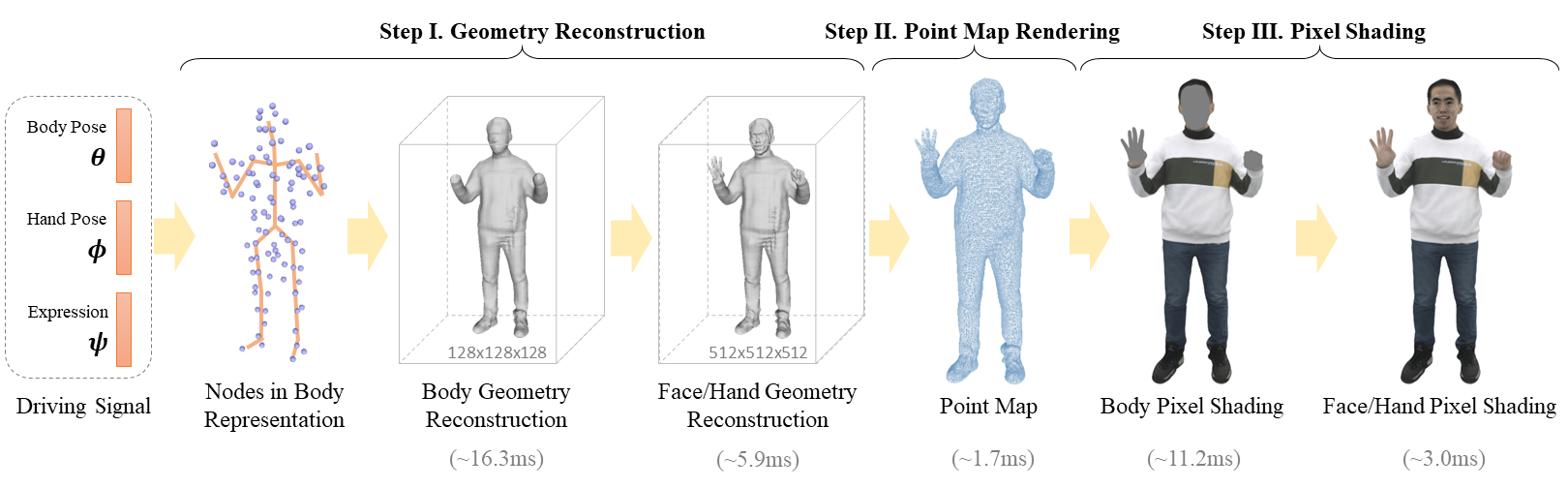}
    \caption{\textbf{Illustration of our real-time rendering pipeline.} Our rendering pipeline firstly reconstructs the geometry model in the form of an SDF volume, from which we render a point map to query the pixel colors given a specific viewpoint. }
    \label{fig:realtime}
\end{figure*}

\subsection{Composition}
\label{sec:avatar_reprez:composition}

With all the necessary building blocks at hand, we can now introduce how we combine different body parts into an expressive avatar. 
There are two key technical designs that assist us towards this goal. One is the usage of SDF, which has a unified, unambiguous definition across different parts. The other one is that our hand representation and face representation are tightly coupled with their underlying parametric templates, which provide correspondences for us to assemble them with the body. 
Using these correspondences, we pre-compute the transformation between the body space and the hand/face space. We also pre-define the blending weights on the SMPL-X mesh, as shown in Figure~\ref{fig:avatar_reprez:body}. Given a point $\bm{p}$ in the body posed space, we first project it onto the SMPL-X mesh and query the blending weight $\omega$. Without loss of generality, let's say the point falls on the left hand. Then we query its SDF value and color in both the body representation and face representation, which we denote as $s_{\mathrm{B}}$, $\bm{c}_{\mathrm{B}}$, $s_{\mathrm{H}}$ and $\bm{c}_{\mathrm{H}}$. 
The final SDF value and color for $\bm{p}$ is computed as:
\begin{equation}
\label{eqn:avatar_reprez:composition:sdf_blend}
    s = \left\{ 
    \begin{array}{ll}
    \omega s_{\mathrm{H}} + (1-\omega) s_{\mathrm{B}}, & \text{if } -0.05 < s_{\mathrm{H}} < 0.025 \\
    s_{\mathrm{B}},  & \text{otherwise} \\
    \end{array}
    \right. .
\end{equation}
\begin{equation}
\label{eqn:avatar_reprez:composition:color_blend}
    \bm{c} = \left\{ 
    \begin{array}{ll}
    \omega \bm{c}_{\mathrm{H}} + (1-\omega) \bm{c}_{\mathrm{B}}, & \text{if } -0.05 <  s_{\mathrm{H}} < 0.025 \\
    \bm{c}_{\mathrm{B}},  & \text{otherwise} \\
    \end{array}
    \right. .
\end{equation}
In this way, we can obtain the final avatar given some specific body poses, hand poses and facial expressions.

\section{Real-time Animation and Rendering}
\label{sec:realtime}

Since our avatar representation is based on NeRF rather than meshes, it cannot be sent to traditional graphics pipelines for efficient rendering. In fact, efficient rendering of NeRF is a difficult problem on its own, and has attracted significant research interest since the debut of NeRF~\cite{reiser2021kilonerf,InstantNGP2022-na44,Plenoctrees2021-nfds81,Chen2022MobileNeRF,DeVRF2022-nfds32}. However, current NeRF acceleration techniques mostly focus on static scenes or dynamic sequence playback, and rely on specific data structures that cannot be easily adapted for avatar animation. 
Therefore, we design a dedicated deferred surface rendering pipeline for our avatar representation. Our deferred surface rendering consists of three steps, namely geometry reconstruction, point map rendering and pixel shading, as illustrated in Figure~\ref{fig:realtime} and described in the following.

\textbf{Step I. Geometry Reconstruction.} In the first step of our deferred rendering pipeline, we reconstruct the complete geometry model in the form of an SDF volume. Note that this step is agnostic to the viewing camera. For the reason of computational efficiency, we do not directly infer a high-resolution SDF volume using the full geometry field in Equantion~(\ref{eqn:avatar_reprez:composition:sdf_blend}). Instead, we follow a coarse-to-fine scheme: we firstly infer only the body geometry field with a low-resolution volume, and then update the hands and the face in the upsampled one.  

Specifically, we first compute the bounding box from the node positions in the body representation. At the center of the bounding box, we create a volume grid with a resolution of 128$\times$128$\times$128, where the side length of each voxel is 2 cm. Such a resolution is not fine enough for representing thin structures like fingers, but sufficient for other major body parts. Recall that in our body representation in Section~\ref{sec:avatar_reprez:body}, a local network $\mathcal{G}_i$ only influences a small local space around node $i$. Based on this property, we directly collect the voxels that fall into the influence space of each node, and pass them through the corresponding local networks in parallel. Since the influence radiuses of all local networks are identical, the voxel numbers assigned to the local networks are similar, which is beneficial for maximizing parallelism. The outputs of the local networks are then gathered and blended following Equation~(\ref{eqn:avatar_reprez:body:feat_fusion}), and finally decoded to SDF values. 

After that, we upsample the SDF volume by a factor of 4, and refine the hands and the face. To this end, we firstly compute the bounding boxes for both hands and the face, and collect the voxels that locate inside the boxes. Then we query the hand geometry field or the face geometry field for these voxels, depending on the bounding boxes they belong to. The final SDF values of these voxels is obtained using the method described in Section~\ref{sec:avatar_reprez:composition}.

\textbf{Step II. Point Map Rendering.} The second step of our rendering pipeline is to raycast the SDF volume to extract views of the implicit surface~\cite{izadi2011kinectfusion}. Given the camera origin and viewing direction, the raycasting operation traverses the SDF volume along camera rays and extracts the intersection between the rays and the zero-crossing surface for each pixel. The intersection point is then recorded as a 2D point map for next step.

\textbf{Step III. Pixel Shading.} In the final step of our rendering pipeline, we compute the pixel colors of the point map to obtain the final rendering results. Similar to geometry reconstruction, we first query the body color field for all the points in the point map and utilize the ``local influence'' property of the local MLPs to parallelize network queries. After that, we query the face color field and the hand color field to update the pixels of the hands and the face, as described in Section~\ref{sec:avatar_reprez:composition}. This produces the final image that corresponds to the given body pose, hand pose, facial expression and view point, marking the end of our rendering pipeline.

Most NeRF-based methods synthesize images through \textit{volume rendering}, which samples millions of points along camera rays for network evaluation. 
In contrast, we design a \textit{deferred surface rendering} pipeline by taking advantage of the implicit surface definition and the disentanglement of geometry and appearance~\cite{sun2022neuconw}. 
Our pipeline first stores geometry reconstruction in an explicit grid and then use it to cull unnecessary points for color inference as much as possible. In this way, we significantly speed up the rendering process of our expressive avatar and finally achieve real-time performance using custom CUDA kernels and modern inference engines like NVIDIA TensorRT. 
Note that this is only possible when the surface prediction is reliable. 
In the next section, we will discuss how we train the network from scratch to acquire the surface, and how we use surface rendering to boost network learning.

\section{Model Training}
\label{sec:training}

In this section, we provide technical details about how we train our expressive avatars. We first present our novel two-pass training strategy in Section~\ref{sec:training:training}.  We then   shortly describe our capture setup and the data pre-processing pipeline, followed by the full training procedure (Section~\ref{sec:training:data})

\subsection{Network Training Strategy}
\label{sec:training:training}

\begin{figure}
    \centering
    \subfigure[Topology-free training.]{
    \includegraphics[width=0.96\linewidth]{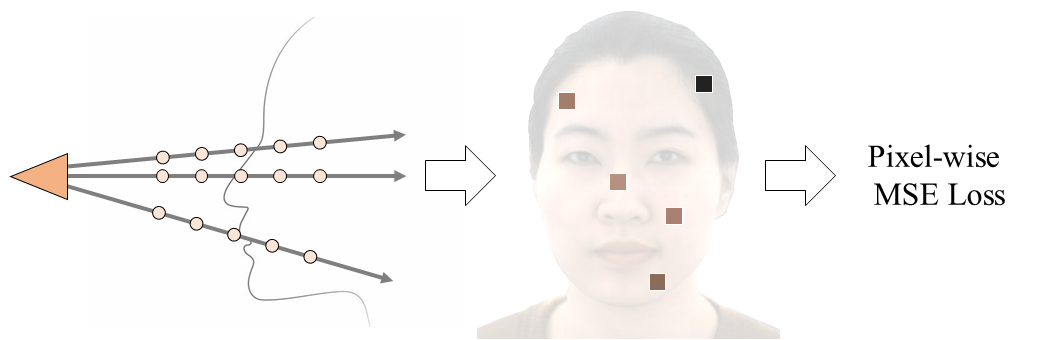}
    }
    \subfigure[Topology-based finetuning.]{
    \includegraphics[width=0.96\linewidth]{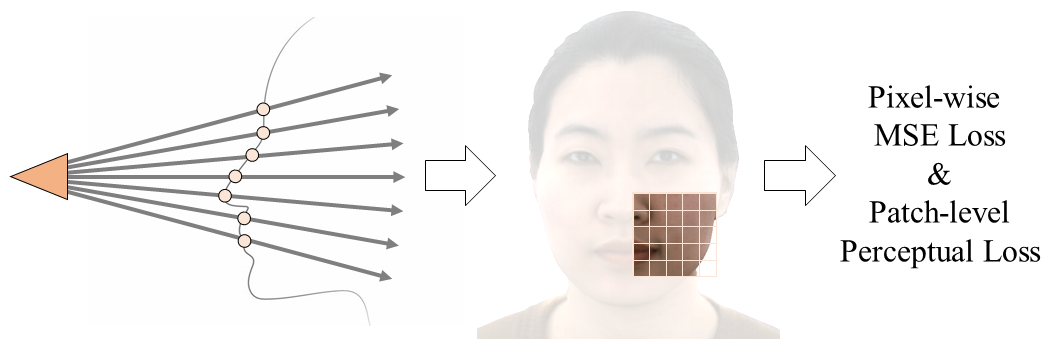}
    }
    \caption{\textbf{Illustration of the two-pass training strategy.}  
    In the topology-free training stage, the networks learn the avatar geometry from scratch with volume rendering on sparse pixels. Then the topology-based finetuning stage utilizes the learned geometry to increase the number of pixel evaluations in one forward pass, allowing us to apply structural supervision with a patch-level perceptual loss.
    }
    \label{fig:twopass_training}
\end{figure}

Similar to other NeRF-based methods, we can train our networks by sampling image pixels, shooting camera rays, performing volume rendering and penalizing the error between the rendered colors and the ground-truth ones~\cite{mildenhall2020nerf}. 
However, we empirically find that the results produced by such a training strategy are not satisfactory enough: the networks trained with this scheme tend to synthesize over-smoothed images and fail to recover dynamic appearance details like the texture patterns on the garments.
This is mainly because volume rendering relies on dense network evaluation along camera rays, which limits the number of pixel queries in one forward pass. Consequently, we could only apply pixel-wise supervision on sparse pixels during network training.   
In addition, this training strategy is based on volume rendering, rather than surface rendering that we use for real-time animation. 
The gap between training and testing further deteriorates the rendering quality of our real-time system. 
To address these issues, we propose a two-pass training strategy. It consists of two stages, namely topology-free training and topology-based finetuning, as illustrated in Figure~\ref{fig:twopass_training} and elaborated in the following. 

\textbf{Pass I. Topology-free training.}
In this stage we assume no prior knowledge about the avatar topology and directly train the full network with volume rendering. 
To this end, we follow the practice of \cite{yariv2021volsdf} and convert the SDF into a density function using:
\begin{equation}
\label{eqn:training:sdf2density}
    \sigma(\bm{p}) = \frac{1}{\gamma} \Phi_{\gamma} (-\mathcal{G}_{*}(\bm{p})), 
\end{equation}
where $\Phi_{\gamma}(\cdot) $ is the Cumulative Distribution Function (CDF) of the Laplace distribution with zero mean, and its scale $\gamma$ is a learnable hyperparameter. 
This transformation links the density in neural radiance fields with our geometry definition, allowing us to optimize the avatar SDF using neural volume rendering.  
At each iteration of network optimization, we randomly fetch a batch of frames, from which we randomly sample a fixed number of pixels to construct the training loss. The loss is defined as:
\begin{equation}
\label{eqn:training:training:loss1}
\begin{split}
    \mathcal{L}_\text{I} = &\mathcal{L}_\text{rgb} + \lambda_\text{mask} \mathcal{L}_\text{mask}  + \lambda_\text{Eikonal} \mathcal{L}_\text{Eikonal} + \\ 
    &\lambda_\text{node} \mathcal{L}_\text{node} + \lambda_\text{ebd} \mathcal{L}_\text{ebd} +  \lambda_\text{KL} \mathcal{L}_\text{KL}, 
\end{split}
\end{equation}
where $\mathcal{L}_\text{rgb}$ measures the MSE between the rendered and true pixel colors, $\mathcal{L}_\text{mask}$ is an MAE loss supervising the occupancy values of the rendered pixels, $\mathcal{L}_\text{Eikonal}$ is the Eikonal loss encouraging the geometry fields to approximate a true signed distance function~\cite{yariv2021volsdf}, $\mathcal{L}_\text{node}$, $\mathcal{L}_\text{ebd}$ and $\mathcal{L}_\text{KL}$ are the regularization losses inherited from \cite{zheng2022structured}. To stabilize network training and obtain consistent performance for different subjects, we do not optimize $\gamma$ in Equation (\ref{eqn:training:sdf2density}) during the training process. Instead, we manually set its value as:
\begin{equation}
\label{eqn:training:training:loss2}
    \gamma (n) = \gamma_{0} + \gamma_{1} \cdot \max \{ 0, (1-n/N) \}, 
\end{equation}
where $n$ is the iteration steps, while $\gamma_{0}=5\times 10^{-4}$, $\gamma_{0}=0.02$ and $N=100000$  are hyperparameters controlling how $\gamma$ varies as training iterates.

\textbf{Pass II. Topology-based finetuning.} 
After the first stage of training, we observe the estimate of the avatar shape is sufficiently reliable although the texture is not clear enough. 
Based on this observation, we design a topology-based finetuning strategy to finetune the color fields in our avatar while keeping the geometry branches fixed. 
Specifically, we sample a set of patches with the size of $H\times H$ on a training frame, and compute the corresponding camera locations and ray directions. Then we leverage ray marching to locate the nearest surface intersection between the rays and the underlying implicit surface, similar to the first two steps in our real-time rendering pipeline. The intersection points are sent to the color networks to query their RGB values, which we take as the pixel color prediction to construct the training loss. Different from the first stage, the training loss in this stage is defined as:
\begin{equation}
    \mathcal{L}_\text{II} = \mathcal{L}_\text{MSE} + \lambda_\text{LPIPS} \mathcal{L}_\text{LPIPS},   
\end{equation}
where $\mathcal{L}_\text{MSE}$ is a simple L2 loss to match pixel-wise appearance with the ground-truth and $\mathcal{L}_\text{LPIPS}$ is a perceptual loss applied on patch level, which provides stronger supervision on high-frequency appearance details as shown in Figure~\ref{fig:exp:eval_vggloss}. We choose VGG as the backbone of LPIPS loss. 

Note that we only apply the training loss on the ray-surface intersection points; we do not back-propagate the loss gradient through the ray marching process. 
Compared to the previous stage where dense point samples are required for pixel color computation, the topology-based finetuning stage predicts pixel colors with a much smaller number of points to be evaluated. This advantage increases the maximum number of pixel evaluations in one forward pass, thus enabling  perceptual supervision on image patches. 
Furthermore, this finetuning stage is built upon surface rendering, similar to the real-time animation system in Section~\ref{sec:realtime}. As a result, it closes the gap between training and testing. 

One may ask why not apply perceptual loss in the first training pass based on volume rendering. In fact, this is possible, as done in HumanNeRF~\cite{weng_humannerf_2022_cvpr}. However, due to the complexity of our network and the memory limits of GPUs, we can only render 400 pixels in one forward pass with volume rendering. Consequently, the patch-level supervision can only be applied on a small patch with a resolution around 20$\times$20, which we found too small to contain any structural information. In contrast, topology-based surface rendering allows us to render a larger patch at 128$\times$128 resolution, enabling stronger supervision on structural accuracy.

\subsection{Data Capture and Processing}
\label{sec:training:data}

\subsubsection{Data Capture} 
\label{sec:training:data:capture}
We use two multi-camera capture systems in this work, one for the full body and the other for the face. The body capture system consists of 16 synchronized cameras, each capable of producing 1500$\times$2048 images. The cameras are evenly distributed around the yaw axis in order to cover the 360-degree view point of the body. For the face capture system, we use 6 synchronized cameras that focus on the frontal face, spanning about 130 degrees horizontally. The image resolution of our face capture system is also 1500$\times$2048. 

We collect data from 4 subjects, two males and one female in two sets of clothing. 
For each subject, we use the full-body capture system to collect video data of some casual motions with neutral face expression, which will be used to train the body avatar and the hand avatar. Each captured sequence is about 2000 frames in length. 
To collect training data for face avatar, we use the face system to capture sequences of 1500-2000 frames in length, which include a range of expressions and natural reading sequences. 
Note that we do not rely on either pre-scanned templates or non-rigid mesh tracking to learn the avatar geometry, which is a departure from existing work~\cite{timur2021driving_signal,Remelli2022TexelAligned,habermann2021realtimeDDC}


\begin{figure}
    \centering
    \includegraphics[width=0.47\linewidth]{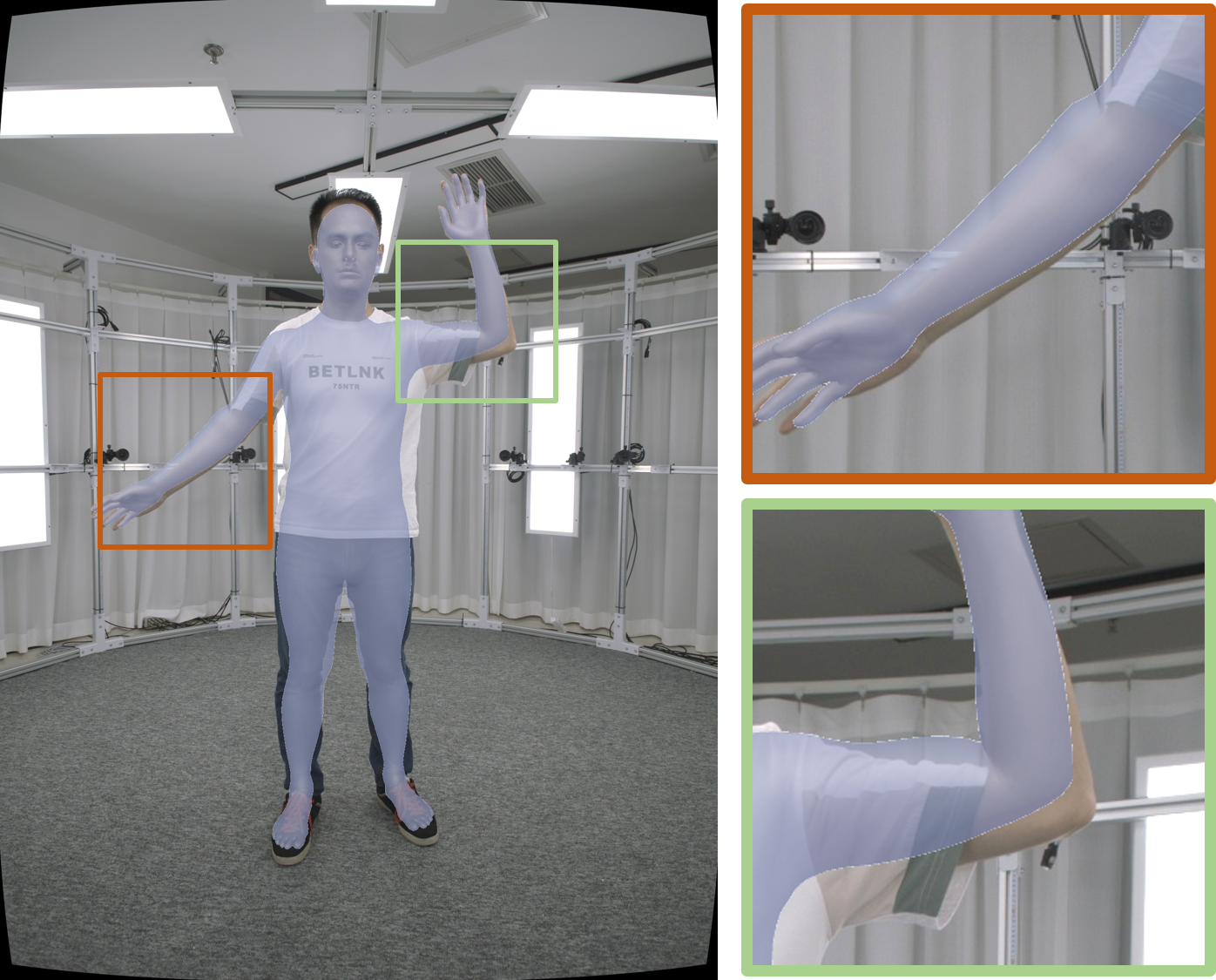}
    \hspace{0.1cm}
    \includegraphics[width=0.47\linewidth]{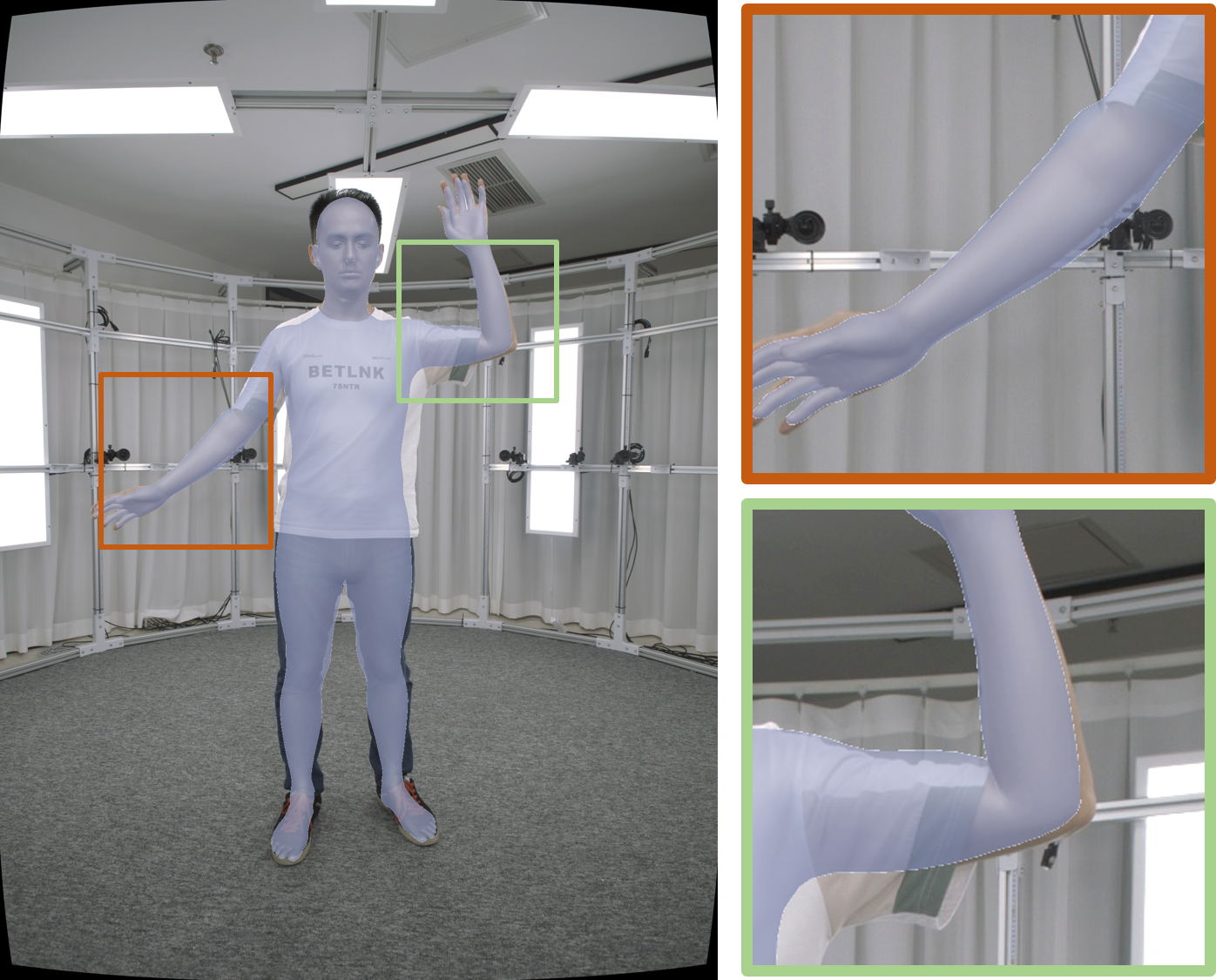}
    \caption{\textbf{Effect of fitting refinement.} Compared to the initial results (left), the refinement step can produce better alignment between the SMPL-X's arms and their image observations (right). }
    \label{fig:eval_smpl_fitting_refinement}
\end{figure}

\subsubsection{Data Pre-processing}
\label{sec:training:data:processing}
To train our networks, we need to obtain SMPL-X fitting for every frame of the full-body sequence and Faceverse fitting for both the full-body sequence and the face sequence. 
For the former purpose, we directly use off-the-shelf tools~\cite{SMPL-X:2019,pymafx2023}. 
However, the fitting results obtained using such a method are not accurate enough due to the sparseness of keypoints and inevitable detection errors. 
Therefore, we regard the results as the initial fitting and design a refinement step based on inverse rendering. 
Specifically, we use Background Matting v2~\cite{BGMv2} to extract the foreground body segmentation and render the silhouette of SMPL-X model with a differentiable rasterizer~\cite{liu2019softras}. 
Then we penalize the inconsistency between them to optimize both the shape and pose parameters for SMPL-X. As shown in Figure~\ref{fig:eval_smpl_fitting_refinement}, the refinement step leads to better alignment between the SMPL-X model and the image observation.  
Please see the appendix for more details.


To register Faceverse, we directly utilize the open-sourced tool provided by Wang \etal~\shortcite{wang2022faceverse}. We also apply Background Matting V2~\cite{BGMv2} for foreground segmentation.

\begin{figure}
    \centering
    \subfigure[Without tone mapping.]{
    \includegraphics[width=0.47\linewidth]{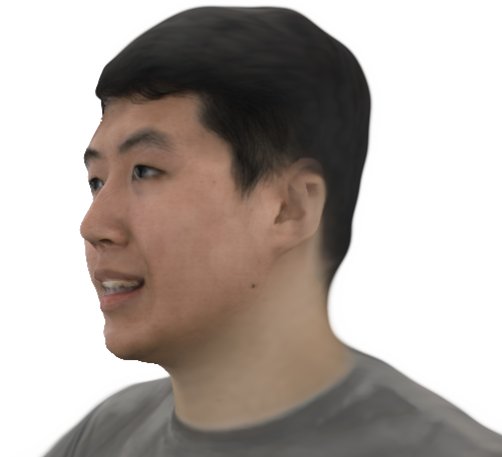}
    }
    \subfigure[With tone mapping.]{
    \includegraphics[width=0.47\linewidth]{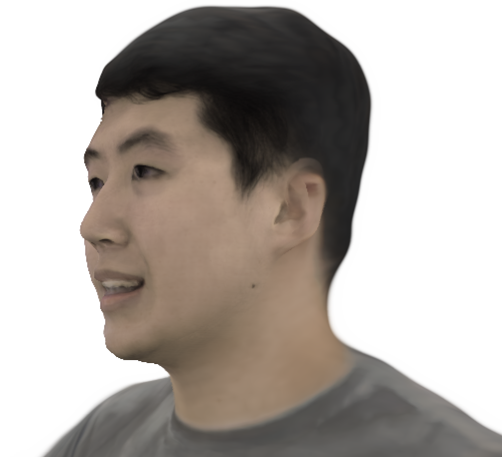}
    }
    \caption{\textbf{Effects of the tone mapping network.} Our tone mapping networks addresses the skin tone difference between body data and face data, and successfully harmonizes the composition results. }
    \label{fig:ablation:tonemapping}
\end{figure}

\begin{figure*}
    \centering
    \includegraphics[width=0.98\linewidth]{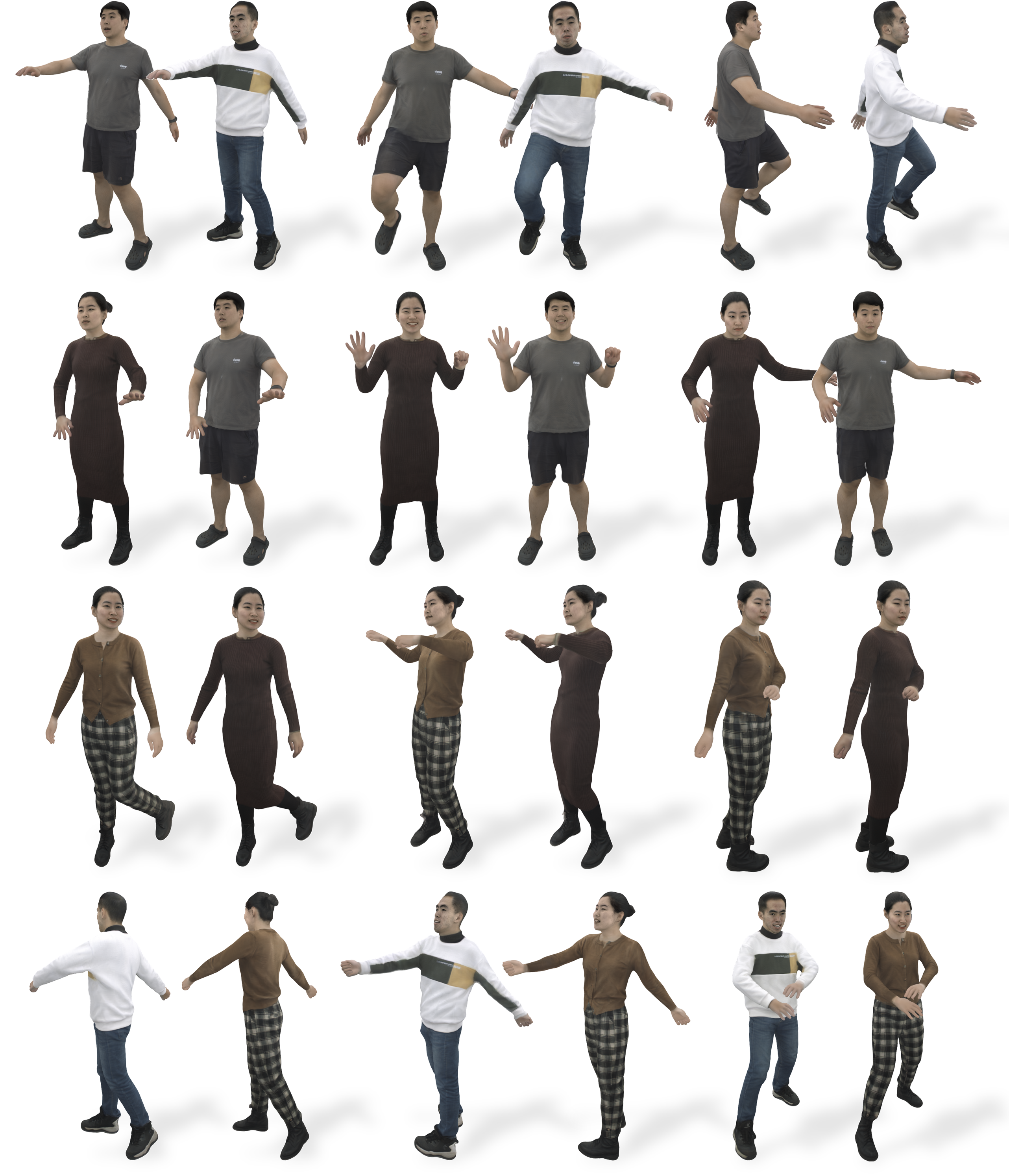}
    \caption{\textbf{Qualitative results on novel pose synthesis.} We train our network for four identities and show the novel pose synthesis results, where two different subjects perform the same motions and expressions. } 
    \label{fig:exp:results}
\end{figure*}

\subsubsection{Full training procedure.}
\label{sec:training:data:full_training}
After collecting the training data for a specific person, we first use the face data \revision{to train} the networks for face representation  with our two-pass training strategy. Then we fix the face networks, and jointly train the body and the hand networks using the body data. This training step also follows the two-pass strategy. To address the color tone difference between the body capture and the facial one, we additionally introduce an tone mapping network to adjust the output of the face color field, as shown in Figure~\ref{fig:ablation:tonemapping}. It is a tiny MLP that takes as input the viewing directions in both the body capture system and the face system, and outputs a mapping matrix $\mathbf{M}\in\mathbb{R}^{3\times 3}$ for tone mapping. The full training pipeline takes about 3 days on one NVIDIA GeForce RTX 3090 GPU with 24 GB GPU Memory, and we report the detailed training statistics in Table~\ref{tab:training}.
Note that the training cost of our method is much lower than existing works that spends two weeks on hundreds of GPUs to build a full-body avatar~\cite{timur2021driving_signal}.  Please refer to the appendix for more training details. 


\begin{table}
  \centering
  \small
  \caption{\textbf{Training statistics of our method.} We report the number of training iterations, the elapsed training time and the number of network parameters that will be updated for each training step. }
    \begin{tabular}{ccccc}
    \toprule
    Part      & Training Pass       & \#Iter. & Time & \#Param. to Update \\
    \midrule
    \multirow{2}{*}{Face} & Pass I  & 200k & $\sim$14 h     & 1.76M \\
          & Pass II & 100k & $\sim$3 h     & 1.10M \\
    \midrule 
    \multirow{2}{*}{Body+Hands} & Pass I  & 300k & $\sim$48 h     & 18.55M \\
          & Pass II & 100k & $\sim$3 h    & 17.57M \\
    \bottomrule
    \end{tabular}%
  \label{tab:training}%
\end{table}%

\begin{figure}
    \centering
    \includegraphics[width=1.0\linewidth]{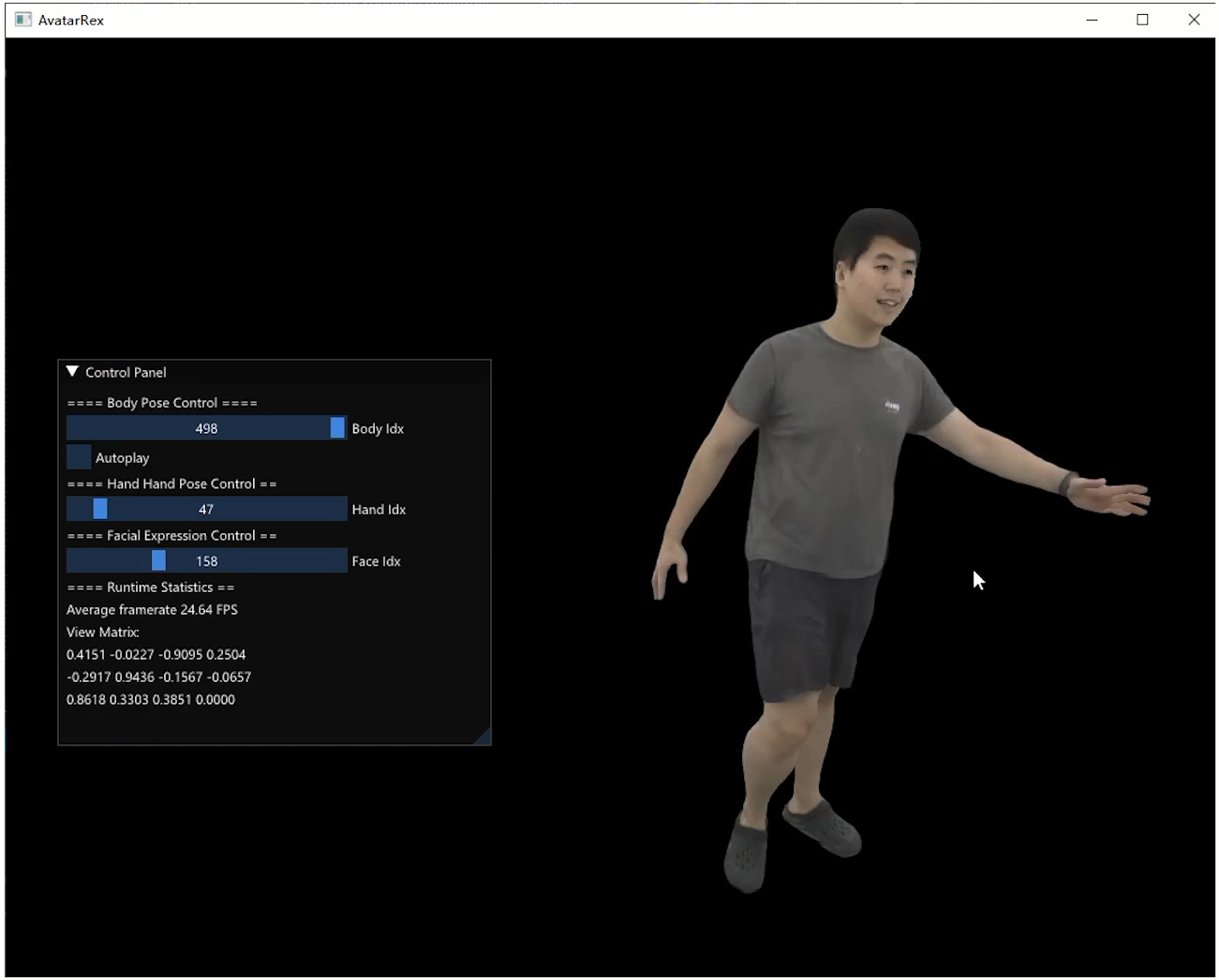}
    \caption{\textbf{Real-time avatar manipulation.} We present an application of interactive avatar manipulation to demonstrate the real-time rendering capability of our avatar.}
    \label{fig:exp:editting}
\end{figure}


\section{Experiments}

This section provides an experimental evaluation of our system for creating expressive full-body avatars that can be rendered \revision{at a real-time framerate}. We first present qualitative results on different identities in Section~\ref{sec:exp:results}, followed by a comparison against state-of-the-art methods in Section~\ref{sec:exp:comparison}. Finally, we study the components of our method in Section~\ref{sec:exp:ablation}. 

\subsection{Results}
\label{sec:exp:results}

As formulated in Section~\ref{sec:avatar_reprez}, the input to our avatar is a set of driving signals consisting of body poses, hand poses and facial expressions. Therefore, we can directly manipulate the pose parameters and expression coefficients to animate our avatar, or in other word, to synthesize novel poses and expressions. In Figure~\ref{fig:teaser} and Figure~\ref{fig:exp:results}, we present some animation examples, where two or more identities follow the same driving signals. The results cover various garment types, cloth materials, body motions and expressions. As we can see from these results, our method supports simultaneous control of the body, the face and the hands together. The results of the male in white long sleeves show that our method can generate photo-realistic dynamic appearance details for different body poses, while the results of another male in short pants show that our method can gracefully handle the relative motion between the legs and the shorts. In addition, we demonstrate the results of a female in different clothing, and prove that our method is more  flexible in modeling different cloth shapes and able to handle the garment type that is not topologically similar to the naked body. 
Furthermore, our avatars can be animated and rendered in real time on a modern high-end graphics card, \textit{e.g.}, 25 FPS at a resolution of 1024$\times$1024 on an NVIDIA GeForce RTX 3090 GPU. To confirm its real-time animation capability, we develop a simple application which allows users to manipulate the body motion, the hand gesture or the facial expression, and the manipulated results can be visualized from any viewpoints in real time, as presented in Figure~\ref{fig:exp:editting}.  
We encourage readers to see our supplemental video for better visualization.

\begin{figure}
    \centering
    \includegraphics[width=1.0\linewidth]{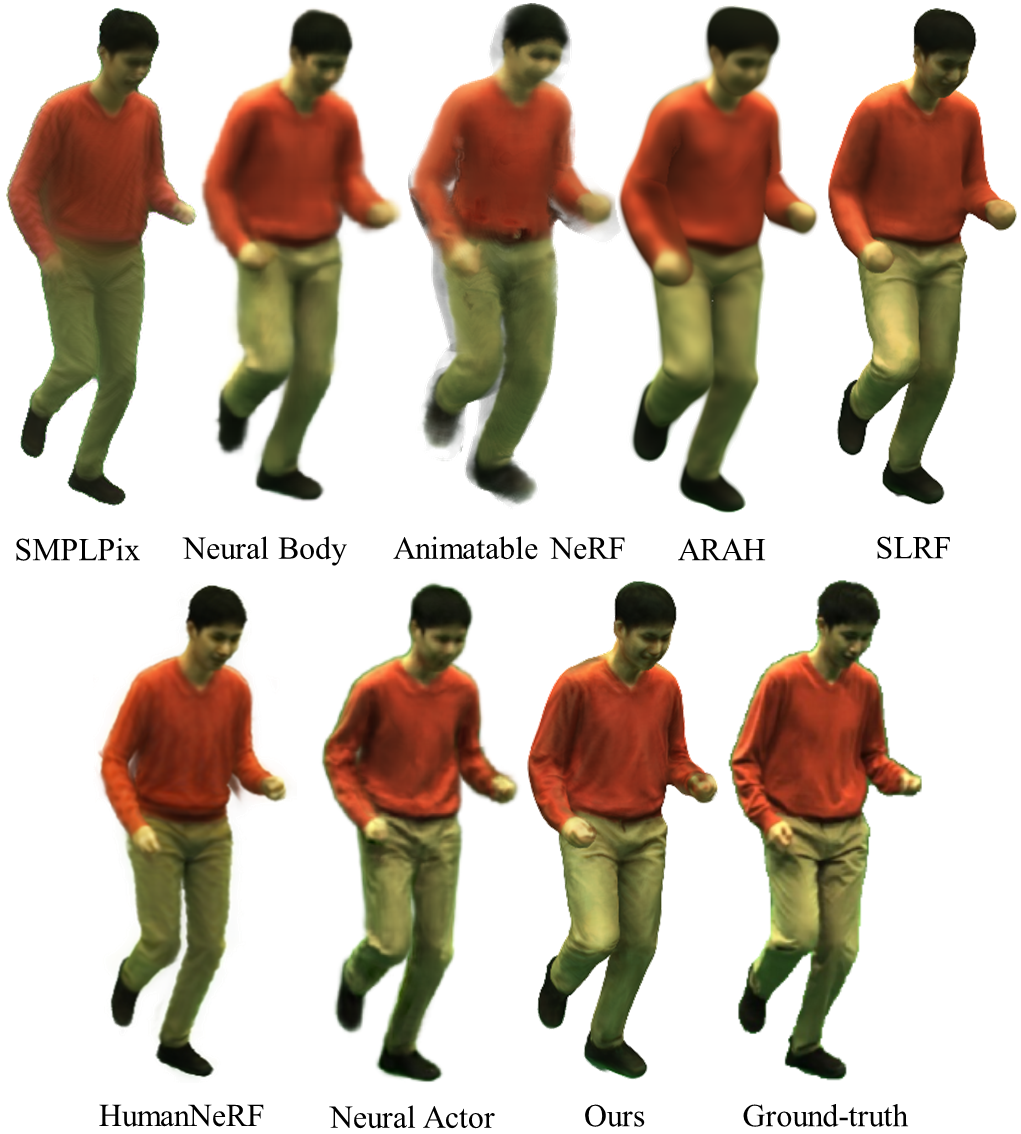}
    \caption{\revision{\textbf{Qualitative comparison against state-of-the-art body avatars.} Given an unseen body pose, our method is able to generate high-quality appearance details. The synthesis quality is comparable with Neural Actor, while outperforming other baselines. }}
    \label{fig:exp:comparison}
\end{figure}

\begin{table}[]
    \centering
    \caption{\revision{\textbf{Quantitative comparison with state-of-the-art body avatars.} To ease reading, we highlight the best scores with orange shading, and the second best with light orange. }}
    \label{tab:comparison}

    \small
    \begin{tabular}{lcccc}
        \toprule
        Methods & PSNR $\uparrow$ & LPIPS $\downarrow$ & FID $\downarrow$ & Framerate $\uparrow$ \\
        \midrule
        SMPLPix             & 23.199 & \cellcolor{orange!20}0.050 & 42.832 & \cellcolor{orange!50}36 fps \\
        Neural Body         & \cellcolor{orange!50}23.946 & 0.096 & 81.527 & 0.32fps \\
        Animatable NeRF     & 22.453 & 0.097 & 76.258 & 0.54fps \\
        ARAH                & 22.070 & 0.092 & 104.330 & 0.07fps \\
        SLRF                & 23.363 & 0.051 & 58.038 & 0.16fps \\
        HumanNeRF           & 23.395 & 0.054 & 39.014 & 0.14fps \\
        Neural Actor        & 23.531 & 0.066 & \cellcolor{orange!50}19.714 & 0.25fps \\
        Ours                & \cellcolor{orange!20}23.709 & \cellcolor{orange!50}0.044 & \cellcolor{orange!20}30.860 & \cellcolor{orange!20}25 fps \\
        \bottomrule
    \end{tabular}
\end{table}

\subsection{Comparison}
\label{sec:exp:comparison}

To validate our method, we compare with recent state-of-the-arts on novel pose synthesis. Unfortunately, existing baselines only model the clothed body while neglecting other fine-grained parts like the hands and the face. For a fair comparison, we also remove the hands and the face in our representation, leaving the body only for comparison. We mainly compared with the following methods:
\begin{itemize}
    \item \textbf{Neural Body}~\cite{peng2021neuralbody}. Neural Body attaches learnable latent codes to the vertices of SMPL model, and employs sparse 3D convolutions to diffuse the latent codes into a radiance field in the 3D space. 
    \item \textbf{Animatable NeRF}~\cite{peng2021animatable_nerf}. Animatable NeRF factorizes a deforming human body into a canonical radiance field and a deformation field that establishes correspondences between the observations and the canonical space. The deformation field is generated from the backward skinning motion of the underlying SMPL model. 
    \item \revision{\textbf{HumanNeRF}~\cite{weng_humannerf_2022_cvpr}. HumanNeRF follows a similar scheme to Animatable NeRF, and introduces an additional non-rigid motion field to better handle large motions like dancing. Additionally, HumanNeRF also introduce perceptual supervision on image patches to facilitate appearance learning.  }
    \item \textbf{ARAH}~\cite{Wang2022ARAH}. ARAH also models a dynamic humans with a canonical field and a motion field, but it uses forward LBS root finding to model the motion field.  
    \item \textbf{SLRF}~\cite{zheng2022structured}. SLRF models the clothed body with a set of structured local radiance fields, which are loosely attached to the SMPL model.  It is the most related work to ours; in fact, our body representation is built upon it. Compared to SLRF, our method not only introduces dynamic feature patches to local fields, but also disentangles geometry and appearance. Both modification are of significant importance for learning high-frequency appearance details.    
    \item \textbf{Neural Actor}~\cite{neural_actors}. Similar to Animatable NeRF and ARAH, Neural Actor also use a canonical radiance field, but it encodes appearance features on the 2D texture maps of the SMPL model to better capture dynamic details. 
    \item \revision{\textbf{SMPLpix}~\cite{Prokudin2021SMPLpix}. Unlike the baselines mentioned above, SMPLpix is a 2D technique that uses neural rendering in the image space. It works by rendering the SMPL vertices onto the image plane and subsequently converting the rendered image into a final RGB image using a neural network based on 2D UNet. Despite its simplicity and speed, such a scheme suffers from view inconsistencies and temporal jittering. }
\end{itemize}

The dataset we use for comparison is the ``Lan'' sequence from DeepCap~\cite{habermann2020deepcap}. It has 33,605 training frames and 23,062
testing frames captured from 11 cameras, covering a large variety of body motions. The resolution of its image frames is 1024$\times$1024. We use the original training/testing split in the dataset for evaluation, and follow the same protocol in \cite{neural_actors}. 
The results of Neural Actors are borrowed from \cite{neural_actors}, while the others are evaluated by ourselves using the original training settings in their implementation.

\begin{figure}
    \centering
    \includegraphics[width=1.0\linewidth]{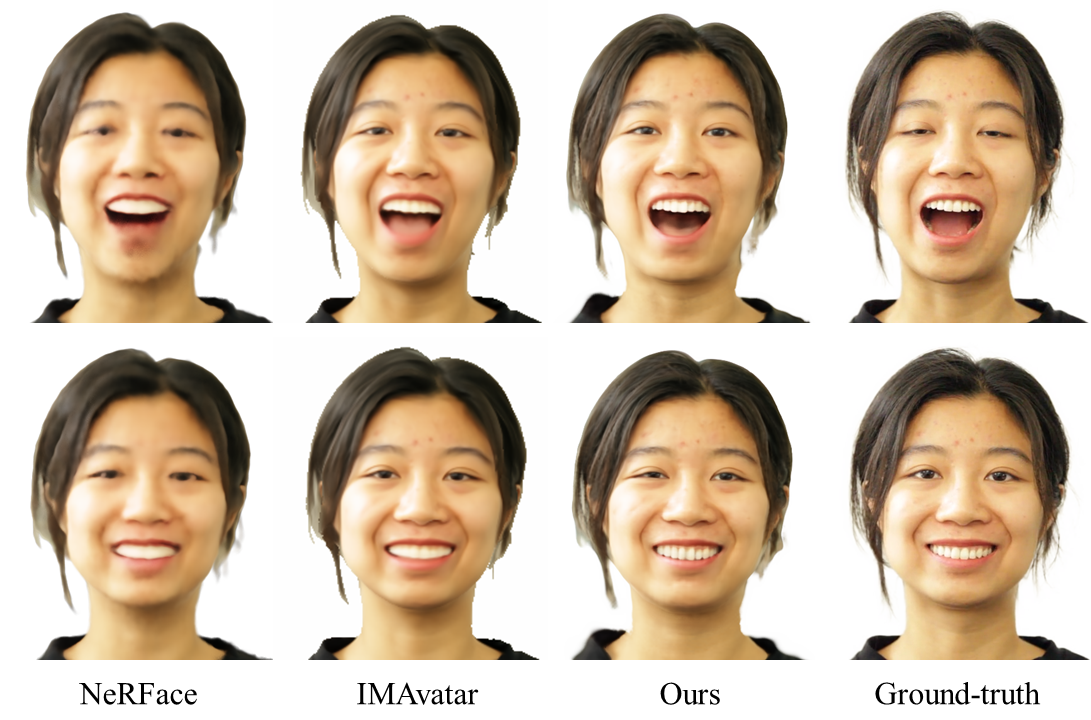}
    \caption{\revision{\textbf{Qualitative comparison against state-of-the-art facial avatars.} Compared to existing methods, ours can generate sharper appearance details like the teeth. }}
    \label{fig:exp:face_comparison}
\end{figure}

\begin{table}[]
    \centering
    \caption{\revision{\textbf{Quantitative comparison with state-of-the-art facial avatars.} To ease reading, we highlight the best scores with orange shading, and the second best with light orange. }}
    \label{tab:face_comparison}

    \small
    \begin{tabular}{lcccc}
        \toprule
        Methods & PSNR $\uparrow$ & LPIPS $\downarrow$ & FID $\downarrow$ \\
        \midrule
        NeRFace  & \cellcolor{orange!20}19.816 & 0.133 & 51.182 \\
        IMAvatar & \cellcolor{orange!50}21.220 & \cellcolor{orange!20}0.092 & \cellcolor{orange!20}46.015 \\
        Ours     & 19.608 & \cellcolor{orange!50}0.089 & \cellcolor{orange!50}33.685 \\
        \bottomrule
    \end{tabular}
\end{table}

\begin{figure}
    \centering
    \includegraphics[width=1.0\linewidth]{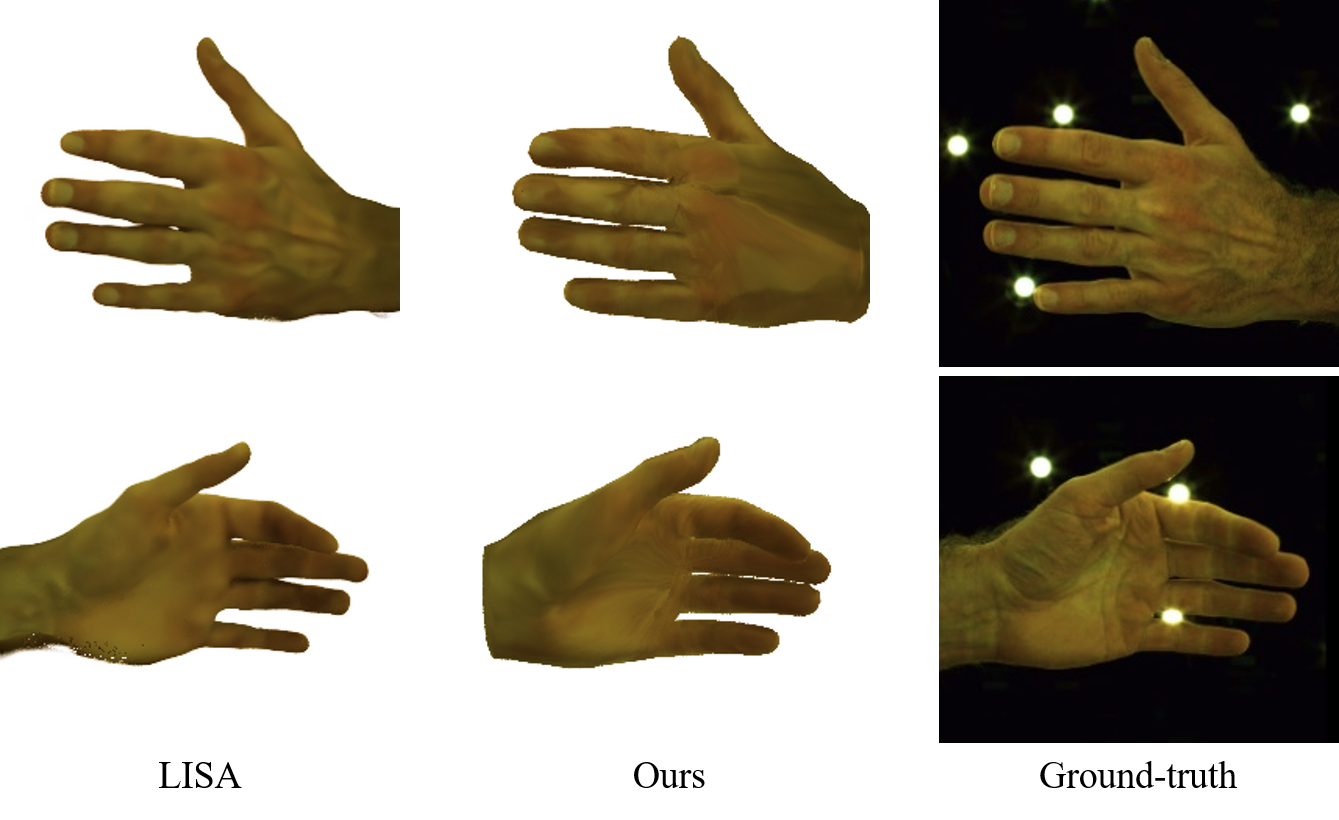}
    \caption{\revision{\textbf{Qualitative comparison against LISA~\cite{Corona2022LISA}.} As shown in the figure, LISA performs better than our method in terms of recovering appearance details like the veins. }}
    \label{fig:exp:hand_comparison}
\end{figure}

\begin{figure*}
    \centering
    \subfigure[Without feature patches or finetuning.]{
    \includegraphics[width=0.3\linewidth]{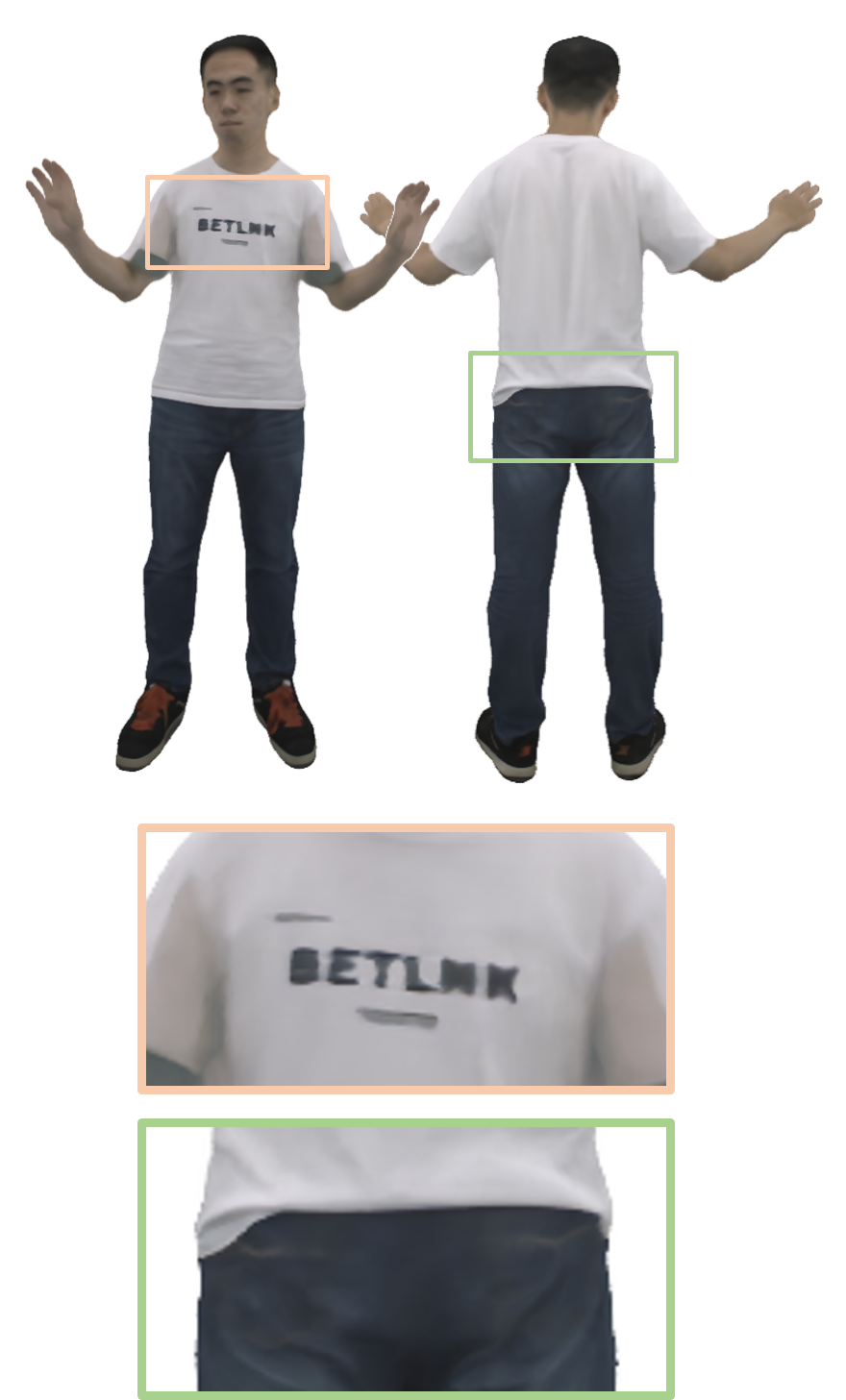}
    }
    \subfigure[With feature patches, without finetuning.]{
    \includegraphics[width=0.3\linewidth]{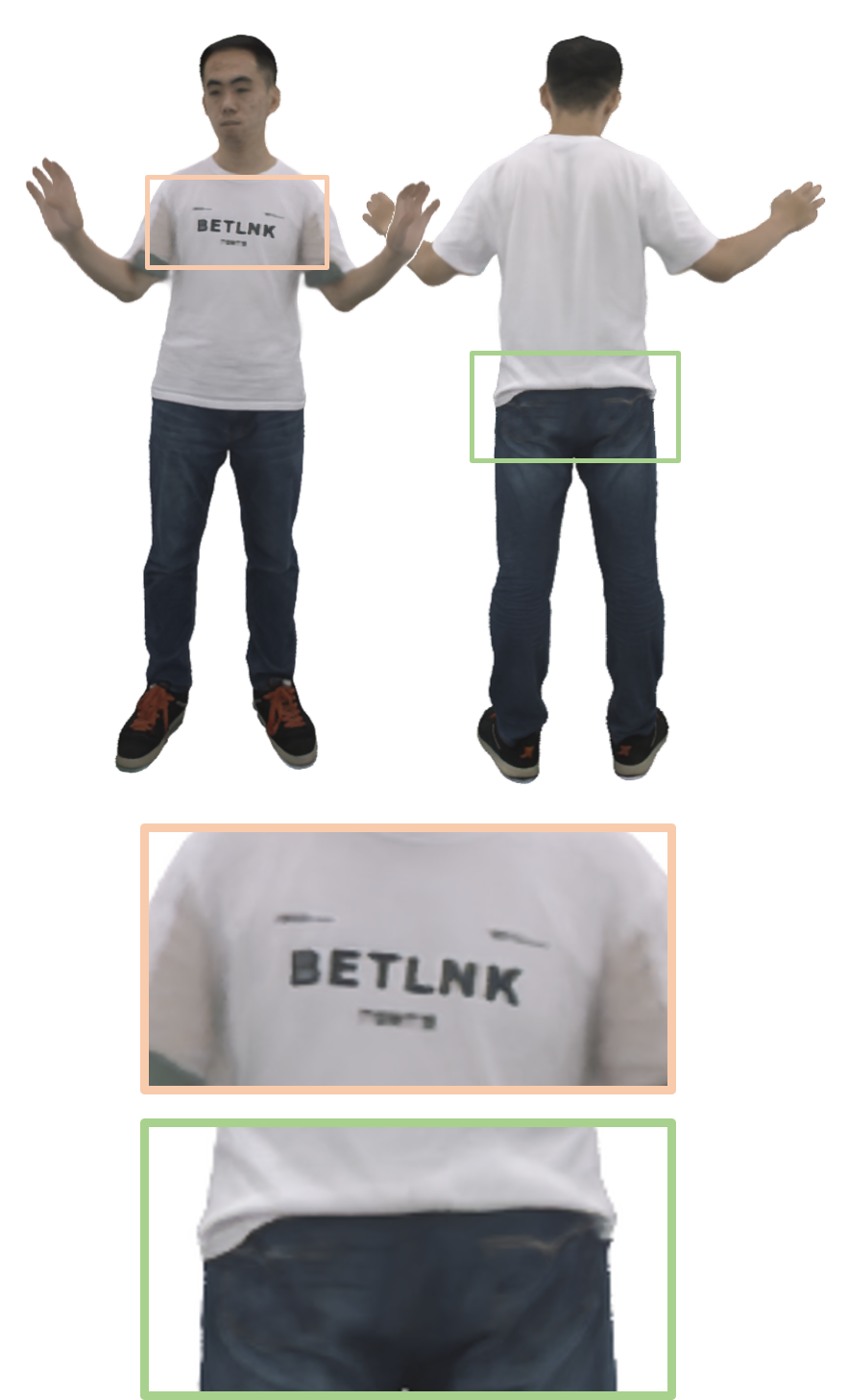}
    }
    \subfigure[Full model.]{
    \includegraphics[width=0.3\linewidth]{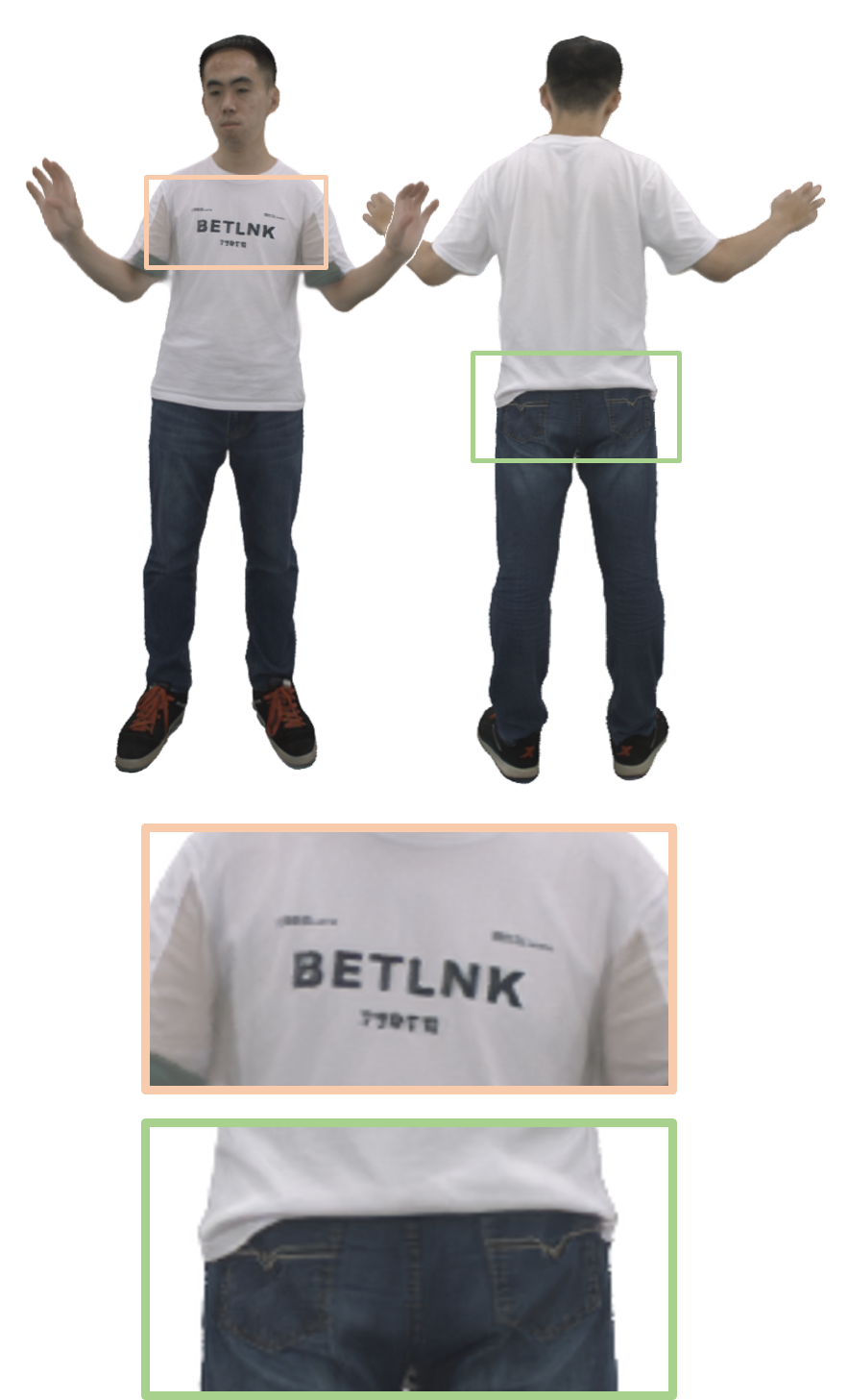}
    }
    \caption{\textbf{Effects of dynamic feature patches and topology-based finetuning on the body model.} Compared to the baseline in (a), assigning a dynamic feature patch leads to stronger capability of detail representation (b), which is further improved by our two-passed training strategy (c).  }
    \label{fig:exp:uvfeat_patchtraining_eval}
\end{figure*}

To measure the quality of novel pose synthesis, we adopt three widely used metrics, namely peak-to-signal ratio (PSNR), learned perceptual image patch similarity (LPIPS)~\cite{Zhang2018LPIPS} and the Frechet inception distance (FID)~\cite{Heusel2017FID}. PSNR simply measures pixel-level accuracy, while LPIPS and FID are more similar to human perception. The numeric results are reported in Table~\ref{tab:comparison}. 
It is easy to see that our method can synthesize image with quality on par to state-of-the-art baselines. 
\revision{
In Figure~\ref{fig:exp:comparison}, we show that our results are comparable with Neural Actor while outperforming other methods in terms of capturing appearance details. This may be attributed to the fact that both Neural Actor and our approach utilize an explicit feature grid to predict high-frequency details, while the remaining methods (Neural Body, Animatable NeRF, ARAH, SLRF, and HumanNeRF) solely rely on MLPs to learn the appearance.  Although the image produced by HumanNeRF shows that perceptual supervision could alleviate this issue, the outcome remains unclear. 
It is worth noting  that Neural Actor performs well only for tight garments due to its reliance on SMPL topology, whereas our local feature patches are more flexible, as we have demonstrated in Section~\ref{sec:exp:results}. Furthermore, our method renders images faster than other NeRF-based methods by two orders of magnitude. 
}



\revision{
Apart from the above comparisons, we also conduct evaluation on our face and hand representations for completeness. 
Specifically, to validate our face representation, we compare with NeRFace~\cite{gafni2021nerface} and IMAvatar~\cite{ZhengABCBH22IMAvatar}, two state-of-the-art baselines on facial avatars. Both baselines and our method are based on implicit representations. Among them, NeRFace~\cite{gafni2021nerface} extends the vanilla NeRF by directly taking the expression coefficients as the additional inputs, while IMAvatar~\cite{ZhengABCBH22IMAvatar} incorporates skinning fields with an implicit representation, leading to better geometry reconstruction and stronger generalization capability. 
The data we use for this experiment is a monocular video sequence released by Zheng \etal~\shortcite{ZhengABCBH22IMAvatar}. It contains 2904 training frames and 1825 testing frames, with a resolution of 512$\times$512. 
We follow the same evaluation protocal as IMAvatar, and use FLAME~\cite{FLAME:SiggraphAsia2017} as the base 3DMM model for both baselines and our method. 
}

\revision{
The qualitative comparison is presented in Figure~\ref{fig:exp:face_comparison}. Although all methods perform well at synthesizing different expressions, our method generates more appearance details such as teeth compared to the baselines. This is due to the feature tri-planes in our method, which provide stronger power for encoding spatially-varying signals. Additionally, our two-pass training strategy encourages the network to fully utilize this advantage.  
The numeric evaluation reported in Table~\ref{tab:face_comparison} confirms that our method is superior to the baselines. 
The results also show that collecting multi-view video is not a necessity for our facial representation; given monocular input, our method can still learn a photo-realistic facial avatar. 
}

\revision{
To evaluate our hand representation, we qualitatively compare with LISA~\cite{Corona2022LISA} on the dataset from InterHand2.6M~\cite{Moon2020InterHand2.6M}. LISA is a state-of-the-art method for hand avatars that models implicit shape and appearance of hands through a collection of rigid parts defined by the hand bones. 
As demonstrated in Figure~\ref{fig:exp:hand_comparison}, LISA outperforms our current approach by recovering more appearance details on the hand skin. However, this comes at the expense of a higher computational load. In the future, we can integrate LISA into our framework to enhance the quality of our hand representation.
}

\subsection{Ablation Study}
\label{sec:exp:ablation}

\begin{table}[]
    \centering
    \caption{\revision{\textbf{Quantitative evaluation of the dynamic feature patches and topology-based finetuning on the body model.} The quantitative results are in line with the findings in Figure~\ref{fig:exp:uvfeat_patchtraining_eval}.}}
    \label{tab:exp:uvfeat_patchtraining_eval}
    \small
    \begin{tabular}{lcccc}
        \toprule
          & PSNR $\uparrow$ & LPIPS $\downarrow$ \\
        \midrule
        Without feature patches or finetuning   & 25.225 & 0.079 \\
        With feature patches, without finetuning   & 25.604 & 0.068 \\
        Full model   & 25.731 & 0.056 \\
        \bottomrule
    \end{tabular}
\end{table}

\begin{figure}
    \centering
    \includegraphics[width=1.0\linewidth]{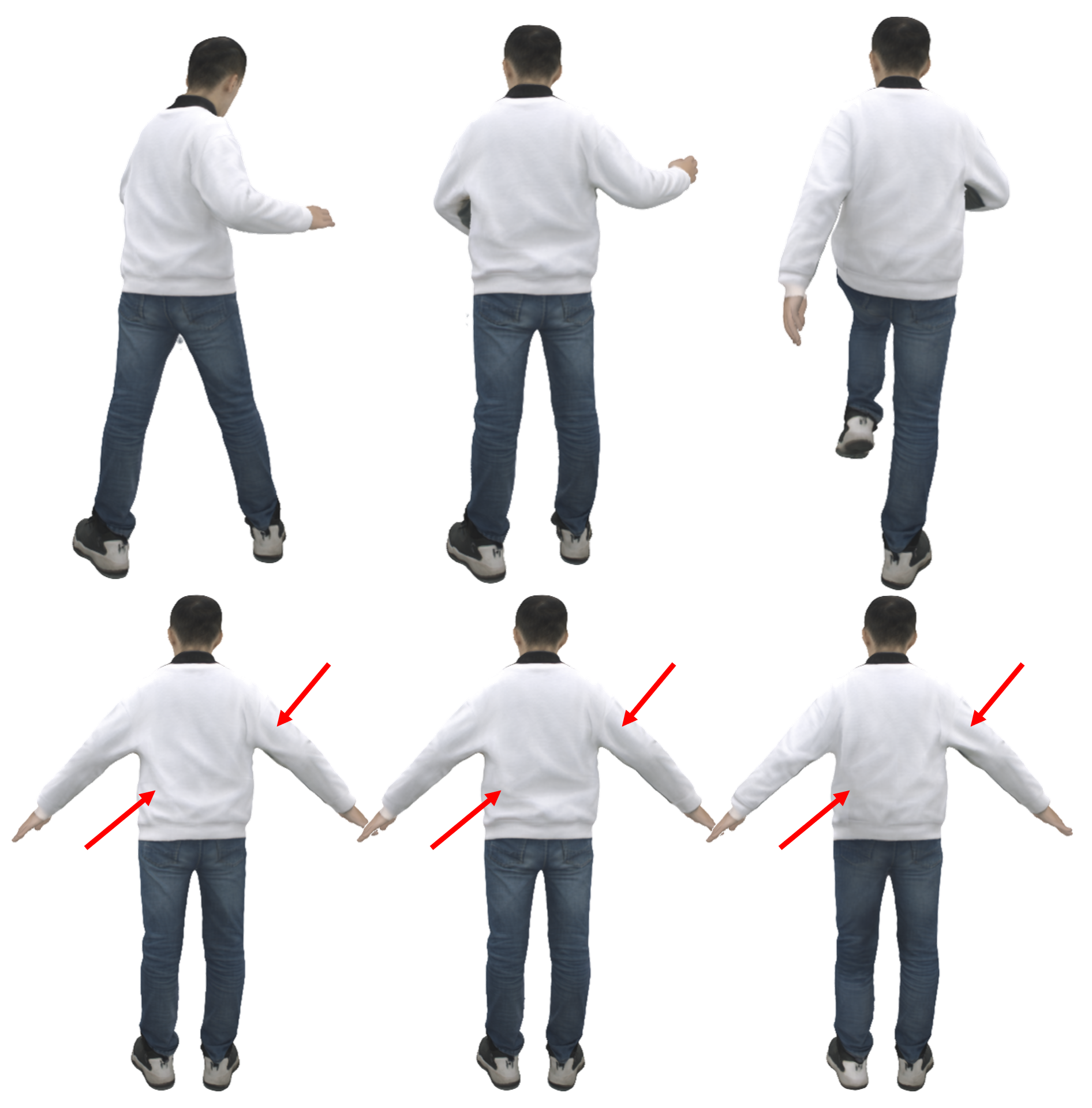}
    \caption{\textbf{Visualization of the effects of dynamic feature patches.} Top: Avatar animation and rendering results. Bottom: avatar rendering with dynamic feature patches under the canonical A-pose. }
    \label{fig:exp:eval_dynauvfeat}
\end{figure}

Here we conduct evaluation on the effects of our novel technical components. First we analyze the effects of the dynamic feature patches in Section~\ref{fig:avatar_reprez:body}. In our method, we introduce a dynamic feature patch for each local color field $\mathcal{C}_i$, and extract feature vectors for the points in the local space of this field. We compare with a baseline that uses purely MLPs to model the local color field without any explicit feature grids, and present the results in Figure~\ref{fig:exp:uvfeat_patchtraining_eval} (a, b) \revision{and Table~\ref{tab:exp:uvfeat_patchtraining_eval}}. We can see that the explicit feature patches allows the network to learn more high-frequency details compared to the baseline. To better understand what the dynamic feature patches encode, we conduct an additional visualization experiment in Figure~\ref{fig:exp:eval_dynauvfeat},  where the avatar are rendered with different feature patches under the same A-pose. As we expect, the feature patches successfully learn the dynamic wrinkles of the garment.

\begin{figure}
    \centering
    \subfigure[Without topology-based finetuning.]{
    \includegraphics[width=0.9\linewidth]{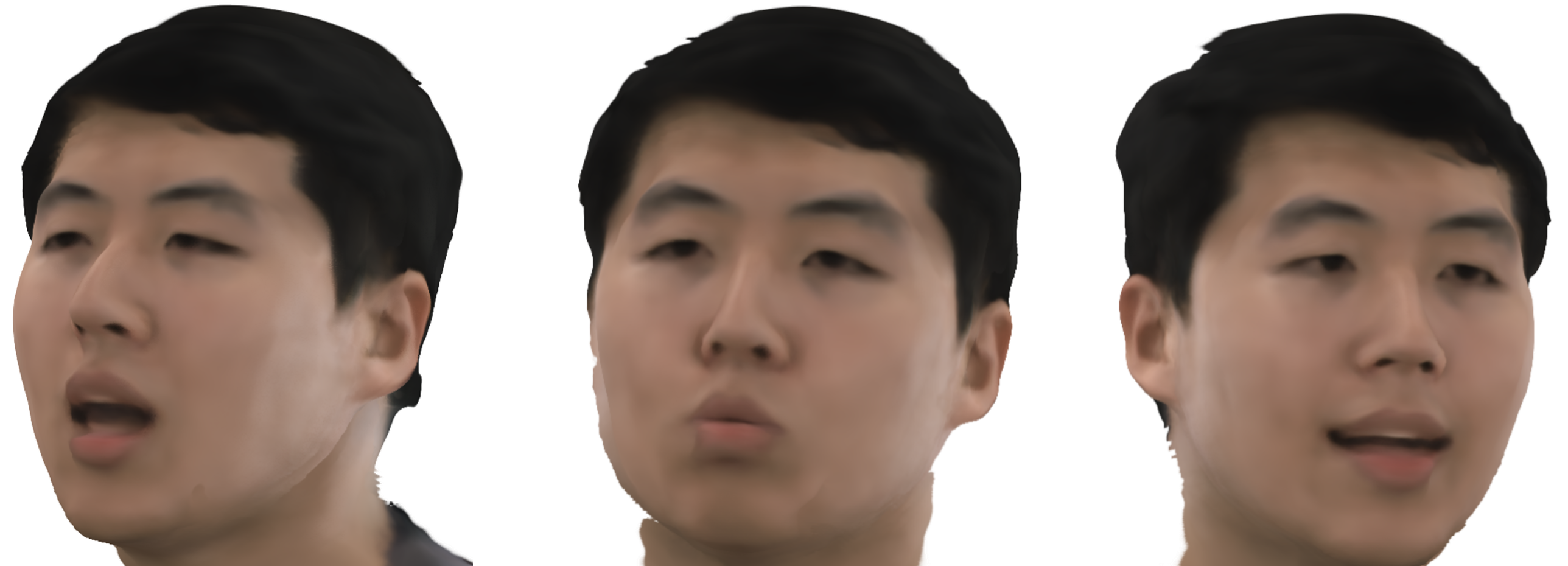}
    }

    \subfigure[With topology-based finetuning.]{
    \includegraphics[width=0.9\linewidth]{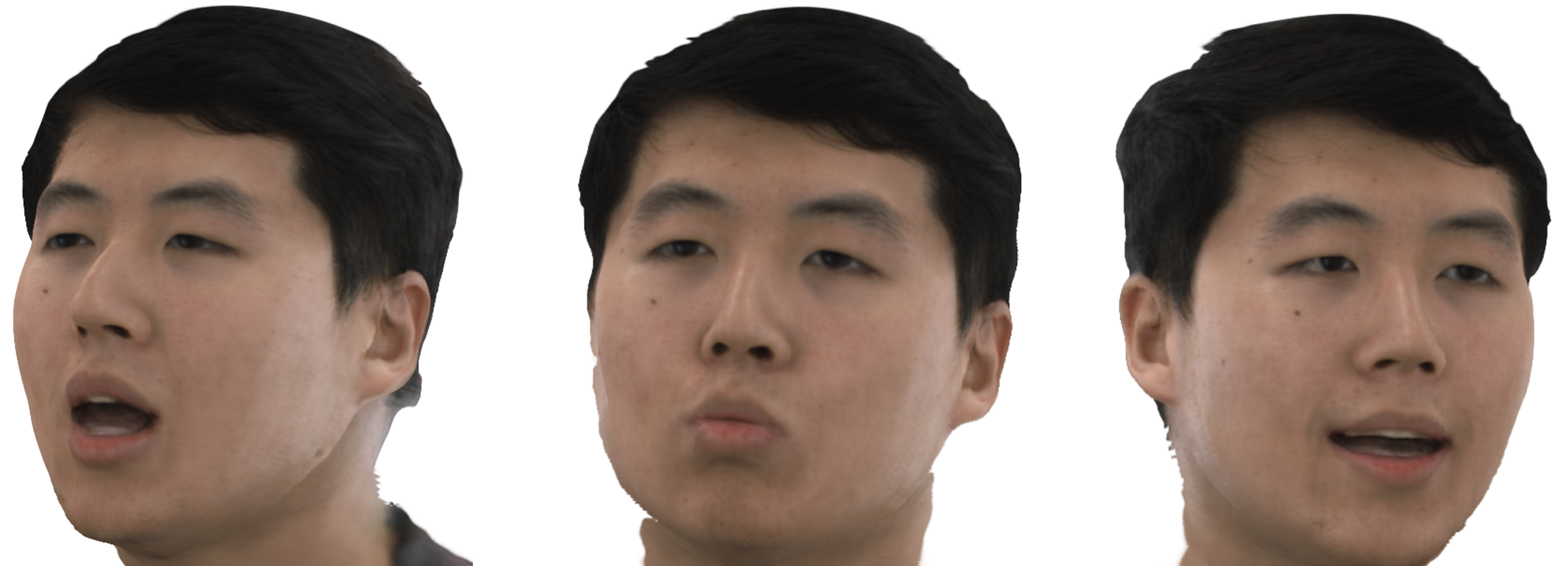}
    }
    \caption{\textbf{Effects of topology-based finetuning on the facial area.} The two-pass training strategy force the face network to learn more photo-realistic facial appearance. }
    \label{fig:exp:eval_face_finetuning}
\end{figure}

\begin{table}[]
    \centering
    \caption{\revision{\textbf{Quantitative evaluation of topology-based finetuning on the facial area.} The quantitative results further confirm that topology-based finetuning leads to perceptually better rendering.}}
    \label{tab:exp:eval_face_finetuning}

    \small
    \begin{tabular}{lcccc}
        \toprule
          & PSNR $\uparrow$ & LPIPS $\downarrow$ \\
        \midrule
        Before finetuning   & 21.109 & 0.202 \\
        After finetuning   & 21.197 & 0.137 \\
        \bottomrule
    \end{tabular}
\end{table}

Next we validate the effectiveness of our two-pass training strategy. In this training strategy, the first pass is the vanilla training process in most NeRF-based methods, while the second pass, \textit{i.e.}, topology-based finetuning, is our contribution. To evaluate its effect, we compared with the results with one-pass training only. As shown in Figure~\ref{fig:exp:uvfeat_patchtraining_eval} (b, c), the topology-based finetuning step force the network to learn more photo-realistic appearance details. This is because the reconstruction loss in the first pass, either in the form of $\ell$1 or $\ell$2, only penalizes pixel-wise error and ignores structural quality. In contrast, the patch-level perceptual loss used in the finetuning step is more sensitive to structure errors and matches human perception better. We conduct similar experiments on the facial area in Figure~\ref{fig:exp:eval_face_finetuning} \revision{and Table~\ref{tab:exp:eval_face_finetuning}}, which further proves the effectiveness of our two-pass training strategy.  

\begin{figure}
    \centering
    \subfigure[Ground-truth.]{
    \includegraphics[width=0.29\linewidth]{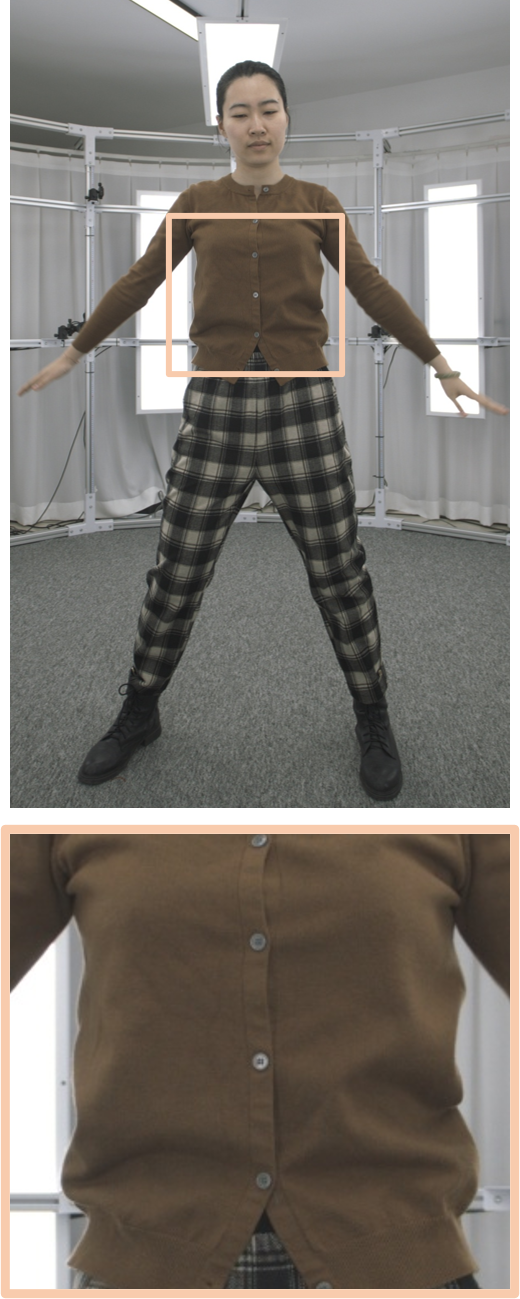}
    }
    \subfigure[Without $\mathcal{L}_\text{LPIPS}$.]{
    \includegraphics[width=0.29\linewidth]{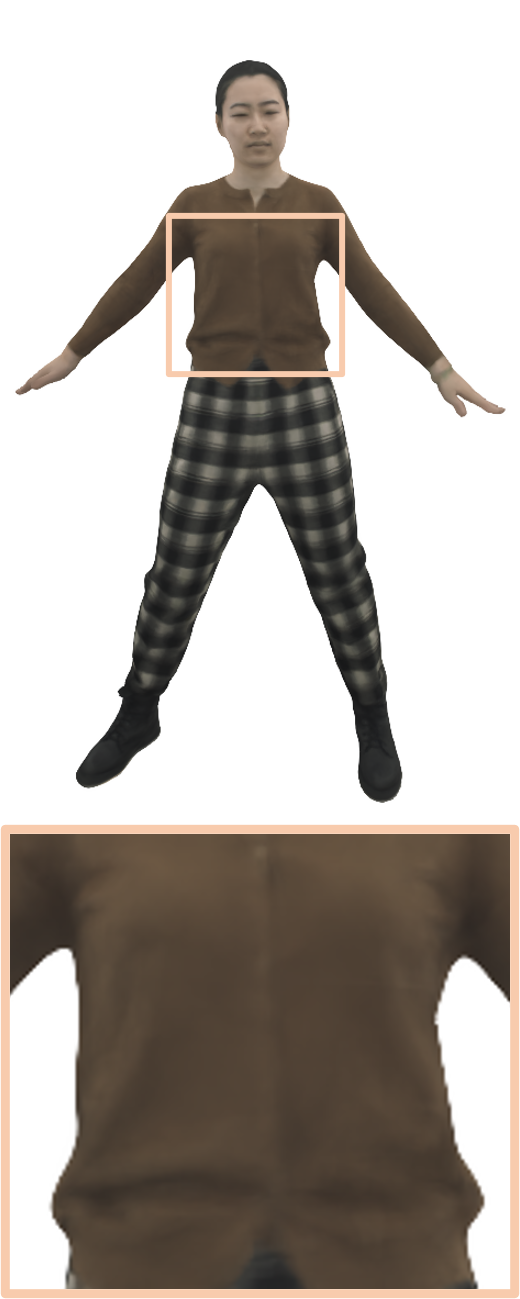}
    }
    \subfigure[With $\mathcal{L}_\text{LPIPS}$.]{
    \includegraphics[width=0.29\linewidth]{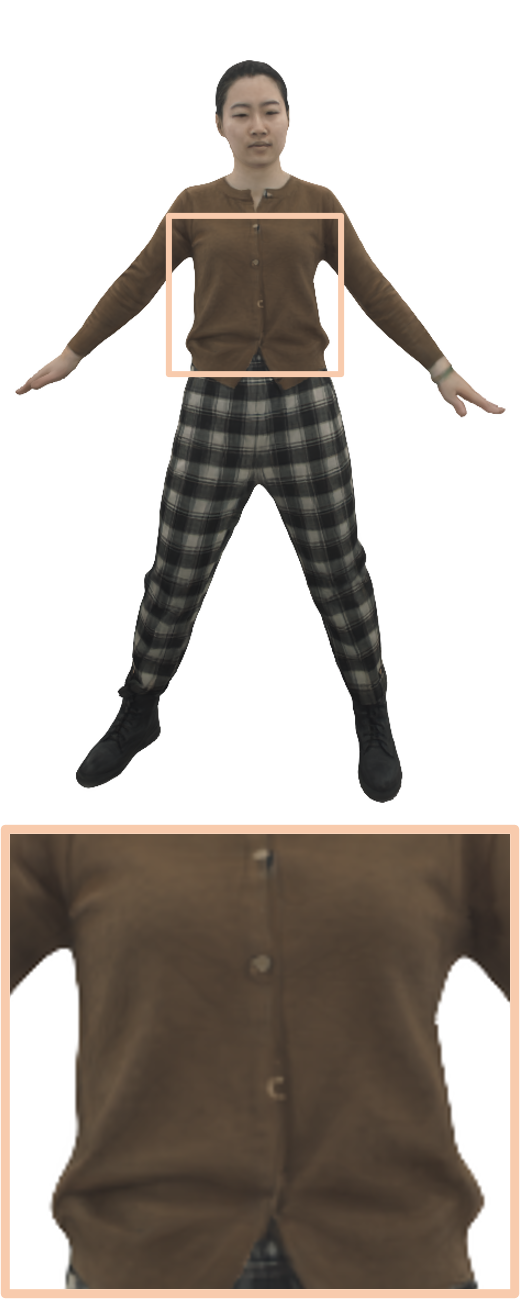}
    }
    \caption{\textbf{Effects of the perceptual loss $\mathcal{L}_\text{LPIPS}$ in topology-based finetuning.} Patch-level supervision with the perceptual loss leads to better recovery of high-frequency details like the wrinkles and the buttons. }
    \label{fig:exp:eval_vggloss}
\end{figure}

Finally we evaluate the role of perceptual loss in the topology-based finetuning pass. We conduct another experiment, where only a simple MSE loss is applied for topology-based finetuning. As presented in Figure~\ref{fig:exp:eval_vggloss}, this experiment confirms again that patch-level supervision is necessary for learning sharp details. Without it, the model only learns blurry appearance. 

\section{Limitation and Future Work}
\label{sec:limitation}

\revision{
Although the results produced by our method is mostly photo-realistic, artifacts occurs occasionally. There are two main factors behind the visual artifacts in our results. Firstly, our method reconstructs a coarse SDF volume as the scaffold for rendering. Its oversmooth nature and inevitable errors result in stitching texture and the artifacts around boundaries when performing ray-casting in the volume. Secondly, our method models the articulated motion around joints through learning the assembling of local fields, instead of relying on full surface of SMPL-X. This may lead to the occasional artifacts around body joints. 
}

Currently our method \revision{uses} one single model to represent the whole clothed human body, no matter how many layers of garments the actor actually wears. Although this is a convenient choice and has been widely adopted in previous work~\cite{timur2021driving_signal,neural_actors,weng_humannerf_2022_cvpr,peng2021animatable_nerf,Wang2022ARAH,zheng2022structured}, it may lead to ghosting effects along the boundary between clothing and skin or between upper and lower garments. 
For our method, these artifacts is mainly caused by the fact that some local fields happen to fall near the garment boundaries and consequently model two cloth layers all together. 
It will result in more noticeable artifacts when the cloth is loose and contains more dynamic deformations. 
For future work, we could replace our unified body representation with a multi-layer one and model different cloth layers separately as in \cite{Xiang2021ModelingClothing}.

Another limitation of our method is about lighting and self-shadowing effects. In particular, when limbs interact with each other or occlude the torso, they cast shadows. However, we do not explicitly address this issue and fully rely on the network to learn the self-shadowing effects as pose-dependent appearance, which adversely impacts generalization. 
In fact, modeling the interactions between lighting and objects in neural radiance fields is still an area of active research~\cite{Srinivasan2021NeRV,Zhang2021NeRFactor,chen2022relighting4d}, and we could employ similar techniques from these works and enable avatar relighting under new illumination.  
Moreover, the SDF fields in our avatar representation also provide a coarse estimation of the underlying avatar geometry, which we could also utilize to approximate the self-shadowing effects as done in \cite{timur2021driving_signal}.

\revision{
Creating full-body avatars involves the modeling of many components, including the body, the face, hands, garments, hairs, eyes, teeth, soft tissues and accessories. This work only considers three dominant parts (i.e., the major body, hands and the face) and is unable to model the components all at once. In addition, the complex dynamics of loose garments, hairs or soft tissues require more sophisticated modeling technology that is beyond the scope of this paper. Therefore, we leave them as future work.}

\revision{
\textbf{Potential Social Impact.} Our method enables automatic creation of a digital "twin" of any individual. This poses a risk of misusing the technology to re-target individuals with poses or actions they never actually perform. To prevent such misuse, it is essential to exercise caution before deploying the technology. Several techniques could be adopted to mitigate this risk. For example, active watermarking of generated content can be employed to detect unauthorized use~\cite{luo2020Distortion}. 
Moreover, forgery detection technology can be utilized to identify manipulated or synthetic imagery in fake videos~\cite{dong2022Protecting,wang2022Domain}. 
By taking these measures, we hope that our technology is used responsibly.
}

\section{Conclusion}
\label{sec:conclusion}

We have presented AvatarReX, a new method for learning full-body avatars from multi-view video data. Compared to existing works, our avatar has two advantages: for one thing, our avatar supports expressive control of the body, hands and the face together; for another, our avatar can be animated and rendered at a real-time framerate with our dedicated rendering pipeline. This is achieved by a compositional representation with the disentanglement of geometry and appearance. Moreover, we introduce a two-pass training strategy that incorporates surface rendering and patch-level perceptual supervision to  further improve the appearance quality. In our experiments, we showcase the capabilities of our method by demonstrating its synthesis results given novel poses and expressions, showing its great potential in many interactive applications.

\begin{acks}
This paper is supported by National Key R\&D Program of China (2022YFF0902200), the NSFC project No.62125107 and No.61827805
\end{acks}

\bibliographystyle{ACM-Reference-Format}
\bibliography{egbib}

\appendix
\section{Network Architecture}
\label{sec:appendix:arch}

\begin{figure}[h]
    \centering
    \subfigure[Body geometry networks. ]{
    \includegraphics[width=0.7\linewidth]{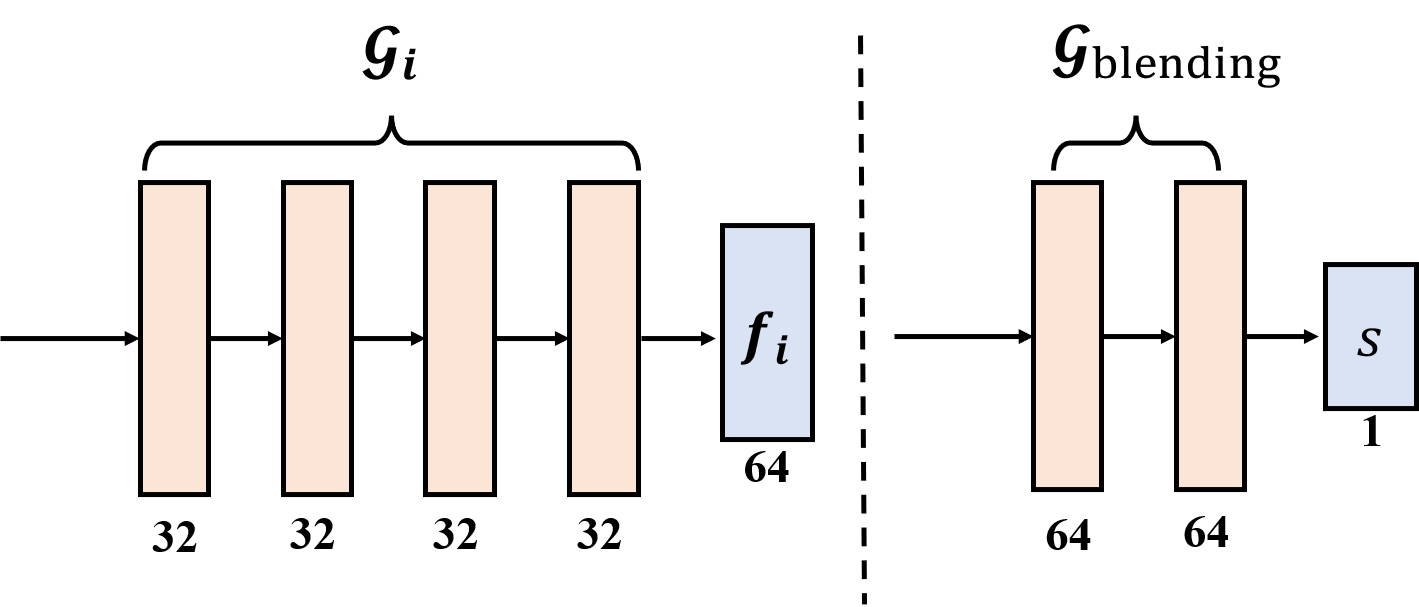}
    }

    \subfigure[Body color networks.]{
    \includegraphics[width=0.7\linewidth]{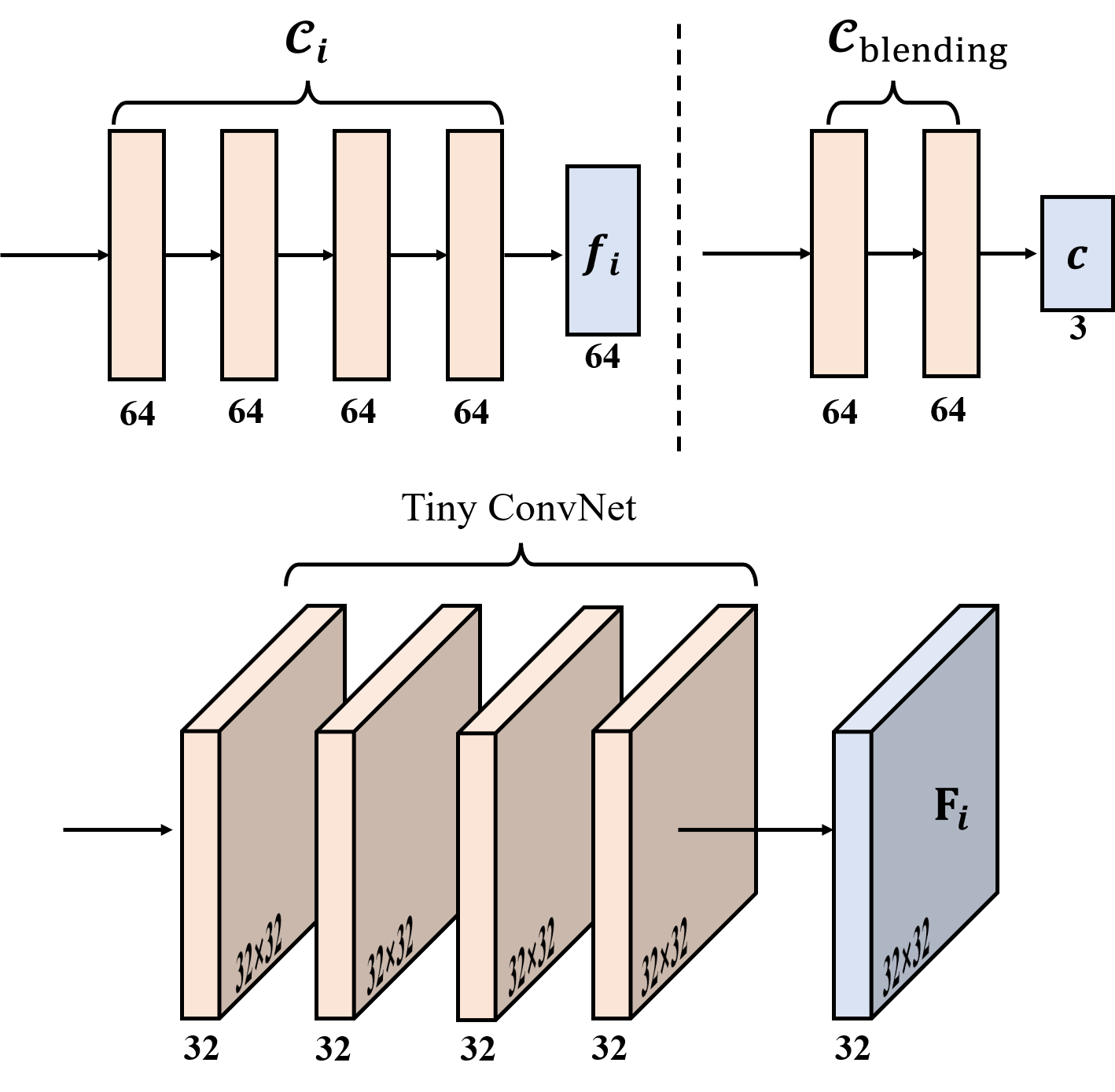}
    }
    \caption{\textbf{Network architecture of our body representation} with the numbers of output channels labeled underneath. All networks are implemented as MLPs except the tiny convolutional network in the bottom of (b). Each fully connected layer in the MLP is followed by ELU activation for the geometry network and ReLU for the color network. For all the layers in the tiny convolutional network, the resolution of the output feature maps is 32$\times$32 and the convolutional kernel size is 3$\times$3. }
    \label{fig:supp:arch_details:body}
\end{figure}

As described in Section~\ref{sec:avatar_reprez:body}, the geometry field of our body representation consists of $N$ local MLPs and a blending MLP with $N$ being the node number, while the color field consists of $N$ local MLPs, a blending MLP as well as $N$ tiny convolutional networks that extract dynamic feature patches from the learnable positional encoding and dynamic detail embeddings. Throughout the paper we sample $N=128$ nodes to construct the structured local representation .We illustrate the body network architecture in Figure~\ref{fig:supp:arch_details:body}. 

\begin{figure}[h]
    \centering
    \includegraphics[width=0.5\linewidth]{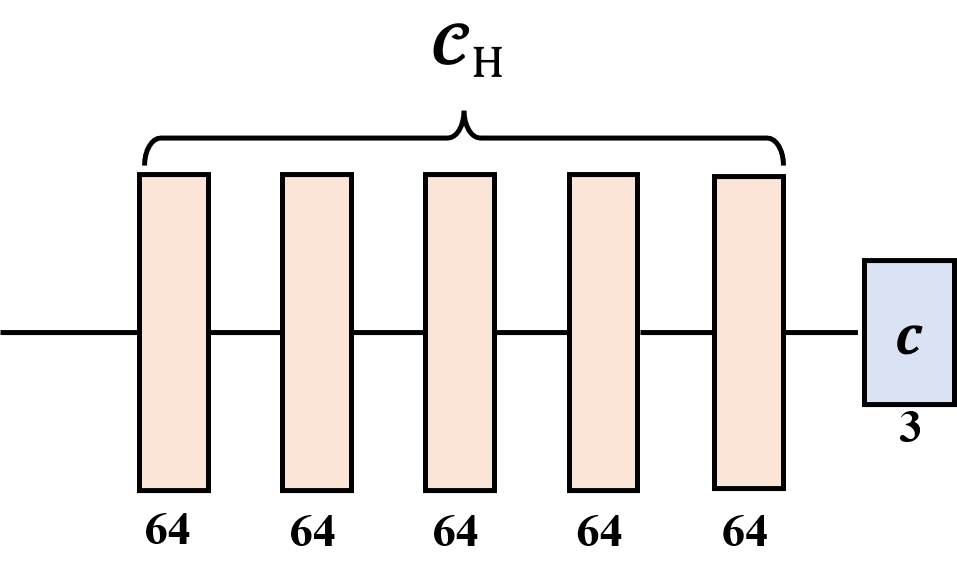}
    \caption{\textbf{Network architecture of our hand color field}  with the numbers of output channels labeled underneath. Each fully-connected layer is followed by ReLU activation. }
    \label{fig:supp:arch_details:hand}
\end{figure}

The hand geometry field in our avatar is derived analytically using Equation~\ref{eqn:avatar_reprez:hand:sdf} , so it does not contain any neural networks. The color field of hands is modeled with a simple MLP, which we illustrate in Figure~\ref{fig:supp:arch_details:hand}.

\begin{figure}[h]
    \centering
    \includegraphics[width=1.0\linewidth]{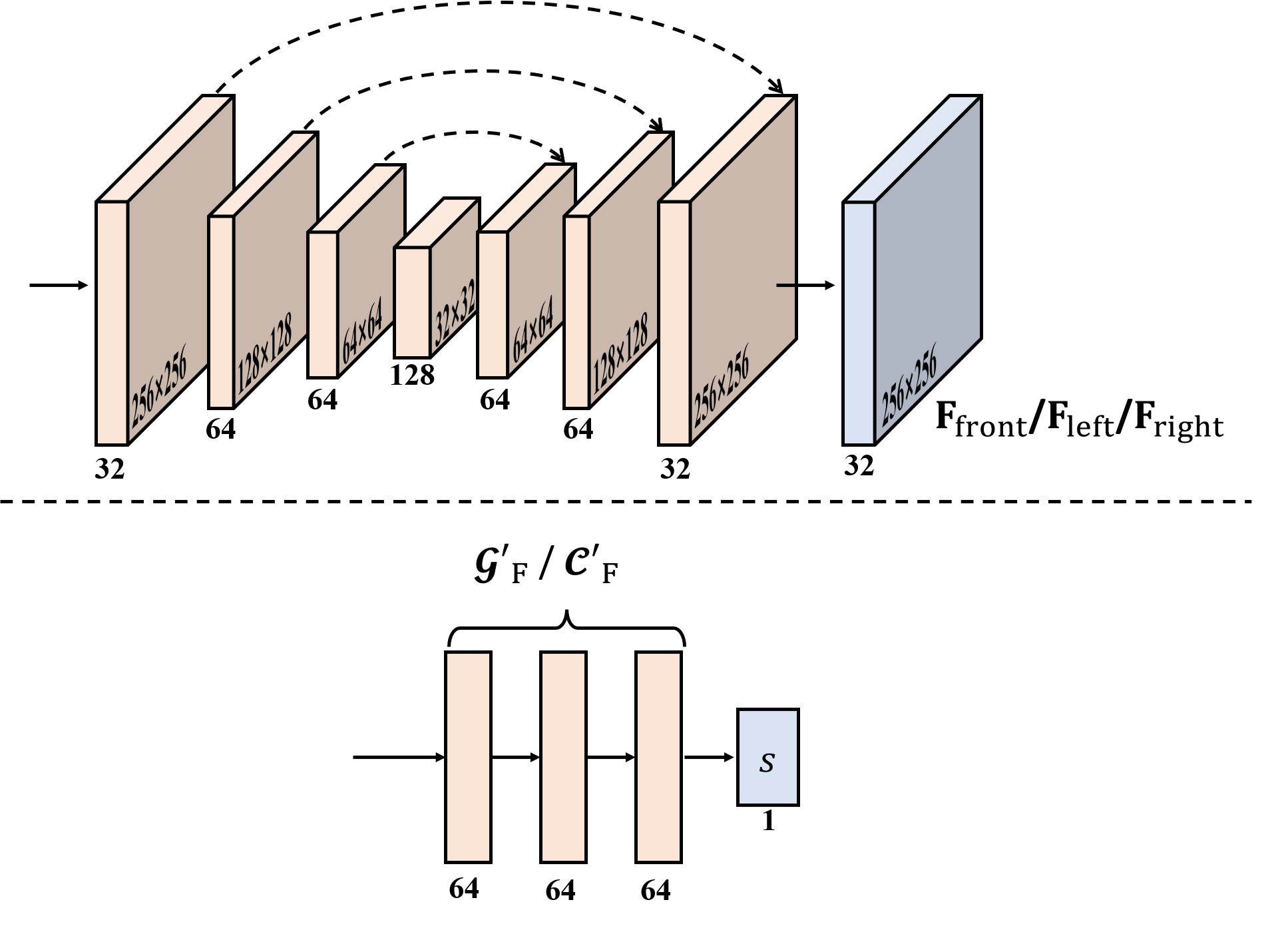}
    \caption{\textbf{Network architecture of our face network}  with the numbers of output channels labeled underneath. Top: the UNet architecture for extracting feature triplanes from orthogonal rendering of Faceverse. Bottom: the MLP architecture for predicting the SDF or color values in the face representation. }
    \label{fig:supp:arch_details:face}
\end{figure}

As shown in Figure~\ref{fig:avatar_reprez:face}, the face network consists of 2 main components, namely a set of UNet for extracting feature triplanes and an MLP to regress the SDF/color value. We use separate sets of networks to model the geometry field and the color field. The detailed network architectures are illustrate in Figure~\ref{fig:supp:arch_details:face}. The resolution of the orthogonal rendering is 256$\times$256.

Note that before feeding the coordinates and view directions into the networks, we augment them using sinusoidal positional encoding~\cite{mildenhall2020nerf}, which is defined as: 
\begin{equation}
    \gamma (\bm{\mathrm{x}}) = \left(\bm{\mathrm{x}}, \sin(\bm{\mathrm{x}}), \cos(\bm{\mathrm{x}}), ..., \sin(2^{m-1}\bm{\mathrm{x}}), \cos(2^{m-1}\bm{\mathrm{x}}) \right). \nonumber
\end{equation}
The value of $m$ is 6 for coordinates and 4 for view directions.

\section{Additional Implantation Details}
\label{sec:appendix:training}

We use PyTorch to implement our networks and use the Adam optimizer to train the networks. The hyperparameters needed for network implementation and training are reported in Table~\ref{tab:training_details}, while the number of iterations is set according to Table~\ref{tab:training}. During network training, the learning rate decays exponentially every 20k iterations.  
Note that the vanilla NeRF adopts a hierarchical sampling strategy, while we only train one network with uniform sampling for volume rendering in the first training pass.  
For baseline methods, we use the author-provided code and run all the experiments using their default training settings.

\begin{table}[h]
    \small
    \centering
    \caption{Hyperparameters for network training and evaluation. }
    \begin{tabular}{lc}
    \toprule
       Parameter Name & Value \\
    \midrule
       $\sigma$ (In Equantion~\ref{eqn:avatar_reprez:body:weight})                         & $0.05$ \\
       $\epsilon$ (In Equantion~\ref{eqn:avatar_reprez:body:weight})                       & $0.001$ \\
       $\lambda_\text{mask}$ (In Equation~\ref{eqn:training:training:loss1})                  & $1.0$ \\
       $\lambda_\text{Eikonal}$ (In Equation~\ref{eqn:training:training:loss1})                  & $2.0$ \\
       $\lambda_\text{node}$ (In Equation~\ref{eqn:training:training:loss1})                  & $0.04$ \\
       $\lambda_\text{ebd}$ (In Equation~\ref{eqn:training:training:loss1})                  & $0.01$ \\
       $\lambda_\text{KL}$ (In Equation~\ref{eqn:training:training:loss1})                  & $1\times10^{-6}$ \\
       $\lambda_\text{LPIPS}$ (In Equation~\ref{eqn:training:training:loss2})                  & $1.0$ \\
       Number of Ray Samples Per Batch (Training Pass I)              & $400$ \\
       Number of Point Samples Per Ray (Training Pass I)             & $64$ \\
       Patch Resolution (Training Pass II)             & $128\times128$ \\
       Batch Size                          & $4$ \\
       Learning Rate  (Training Pass I)                                & $5\times 10^{-4}$  \\
       Learning Rate  (Training Pass II)                                & $1\times 10^{-4}$  \\
    \bottomrule
    \end{tabular}
    \label{tab:training_details}
\end{table}

In order to achieve real-time testing performance, we implement our rendering pipeline fully on GPU using CUDA/C++. The network inference is performed using NVIDIA TensorRT, while the other parts of our rendering pipeline is implemented with custom CUDA kernels. The running time of each step in reported in Figure~\ref{fig:realtime}.

\section{Additional Details of Data}
\label{sec:appendix:data}

\begin{figure}[h]
    \centering
    \includegraphics[width=1.0\linewidth]{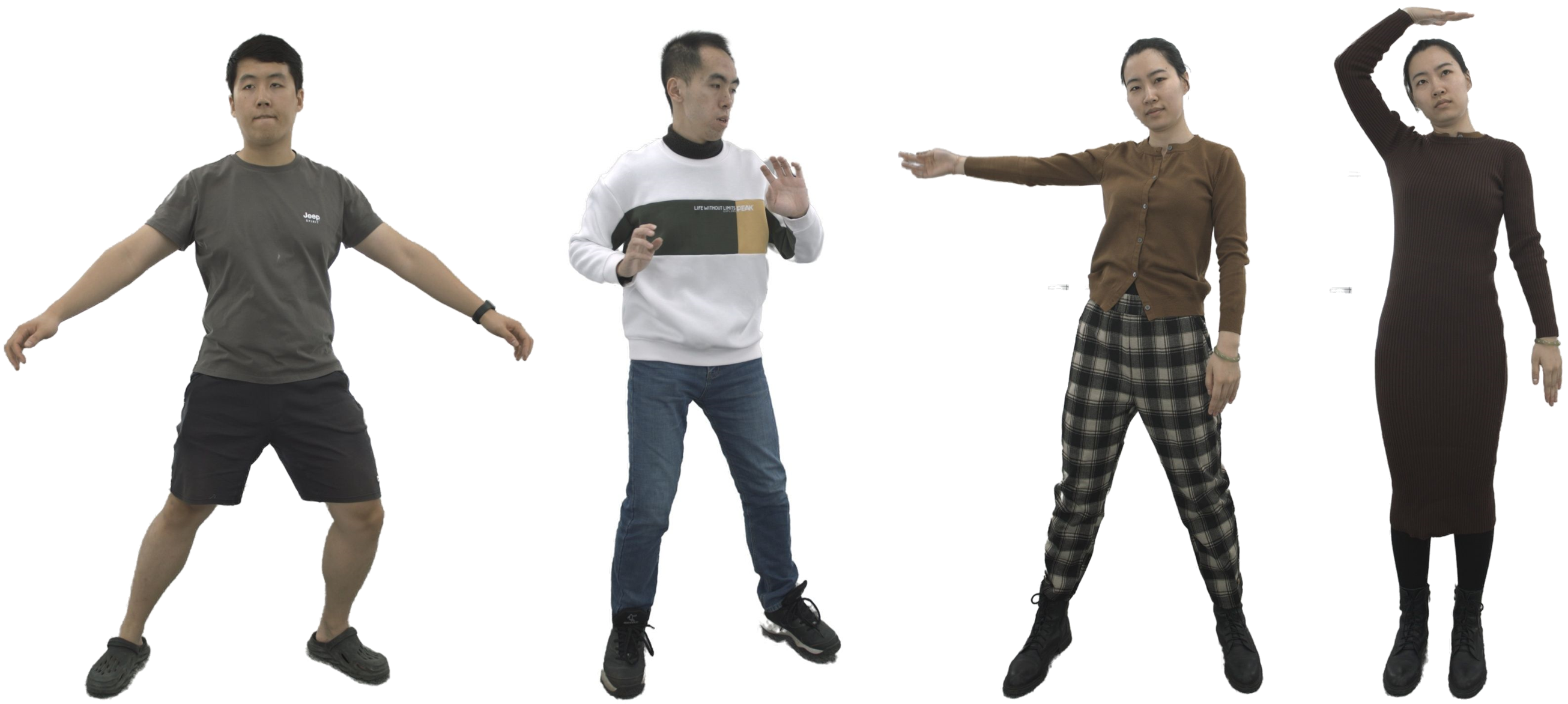}
    \caption{Subjects in our experiments. }
    \label{fig:supp:dataset_example}
\end{figure}

In our experiments, we use the capture system in Section~\ref{sec:training:data:capture} and collect multi-view video data for 4 subjects, two males and one female in two sets of clothing. Figure~\ref{fig:supp:dataset_example} present some example video frames. 

To obtain the initial SMPL-X fitting for training data, we follow a classical optimization-based method~\cite{SMPL-X:2019}. Then we refine the initial fitting results using differentiable rendering. 
Specifically, we use Background Matting v2~\cite{BGMv2} to extract the foreground body segmentation $M_t^{*}$ for frame $t$. Then we render the silhouette of SMPL-X model with a differentiable rasterizer~\cite{liu2019softras}. The rendered silhouette is denoted as $M(\mathbf{\theta}_t, \mathbf{\beta})$, where $\mathbf{\beta}$ are shape coefficients of SMPL-X model and $\mathbf{\theta}_t$ is the pose parameters at frame $t$. We optimize $\mathbf{\theta}_t$ and $\mathbf{\beta}$ through the following energy function: 
\begin{equation}
\label{eqn:training:smplx_fitting_loss}
    \mathcal{E} = \mathcal{E}_{sil} + \lambda_1 \mathcal{E}_{kpt} + \lambda_2 \mathcal{E}_{reg}, 
\end{equation}
where $\mathcal{E}_{sil}$ measures the MSE between $M_t^{*}$ and $M(\mathbf{\theta}_t, \mathbf{\beta})$, $\mathcal{E}_{kpt}$ penalizes keypoint reprojection errors with an L2 loss, and $\mathcal{E}_{reg}$ serves as a regularization term to prevent the parameters from deviating from the initial values. The regularization term $\mathcal{E}_{reg}$ is defined as:
\begin{equation}
\label{eqn:smplx_fitting_loss_reg}
    \mathcal{E}_{reg}=||\bm{\theta}-\bm{\theta}_{init}||_2^2 + ||\bm{\beta}-\bm{\beta}_{init}||_2^2, 
\end{equation}
where $\bm{\theta}_{init}$ and $\bm{\beta}_{init}$ are the initial value of pose parameters and shape coefficients, respectively. 
For all our experiments, we set $\lambda_1=1\times10^{-6}$ and  $\lambda_2=1\times10^{-3}$. 

For computational efficiency, we first optimize $\mathbf{\theta}_t$ and $\mathbf{\beta}$ for 20 frames that are uniformly sampled from the sequence. After that, we fix the shape coefficients and optimize the pose parameters for each individual frame. 

\section{Results on ZJU-Mocap}
\label{sec:appendix:exp}

\revision{
This paper focuses on developing full-body avatars that can be driven with everyday actions such as walking, talking, sports and so on. Therefore, we collect training videos of about 2000 frames in length, capturing common body movements and usual facial expressions. While we recommend users to use our method in this setting, it is worth noting that our method still works with shorter videos and fewer cameras, such as those in the ZJU-Mocap system (which uses only four cameras and has a length of 300 frames). In Table~\ref{tab:supp:comparison_zjumocap}, we report the numeric results of our method's performance with regards to pose generalization on Sequence ``387'' from ZJU-MoCap. The results show that our approach can achieve comparable performance with state-of-the-art methods  under such extremely sparse inputs. 
}

\begin{table}[]
    \centering
    \caption{\revision{\textbf{Quantitative comparison on ZJU-Mocap~\cite{peng2021neuralbody}.} We report PSNR and LPIPS on synthesized images under unseen poses from the testset of the ZJU-MoCap. To ease reading, we highlight the best scores with orange shading, and the second best with light orange. }}
    \label{tab:supp:comparison_zjumocap}

    \small
    \begin{tabular}{lcc}
        \toprule
        Methods & PSNR $\uparrow$ & LPIPS $\downarrow$\\
        \midrule
        Neural Body         & 22.7 & 0.135 \\
        Animatable NeRF     & 23.1 & 0.145 \\
        ARAH                & \cellcolor{orange!50}24.2 & \cellcolor{orange!50}0.099 \\
        SLRF                & \cellcolor{orange!20}23.6 & 0.109 \\
        Ours                & \cellcolor{orange!20}23.6 & \cellcolor{orange!20}0.104 \\
        \bottomrule
    \end{tabular}
\end{table}



\end{document}